\documentclass[lettersize,journal]{IEEEtran}
\usepackage{amsmath,amsfonts}
\usepackage{algorithmic}
\usepackage{algorithm}
\usepackage{array}
\usepackage[caption=false,font=normalsize,labelfont=sf,textfont=sf]{subfig}
\usepackage{textcomp}
\usepackage{stfloats}
\usepackage{url}
\usepackage{verbatim}
\usepackage{graphicx}
\usepackage{cite}
\usepackage{pifont}
\usepackage{adjustbox}
\usepackage{multirow}
\usepackage{booktabs}
\usepackage{bbm, dsfont}
\usepackage{amssymb}
\usepackage{xcolor}
\usepackage{colortbl}
\usepackage{bbding}
\usepackage{fontawesome5}
\usepackage[pagebackref=true,breaklinks=true,colorlinks,bookmarks=false,linkcolor=blue,citecolor=blue,urlcolor=blue]{hyperref}

\begin{document}


\title{Unifying Color and Lightness Correction with View-Adaptive Curve Adjustment for Robust 3D Novel View Synthesis}


\author{Ziteng Cui, Shuhong Liu, Xiaoyu Dong, Xuangeng Chu,  Lin Gu~\Envelope,  Ming-Hsuan Yang, Tatsuya Harada
\thanks{Ziteng Cui, Shuhong Liu, Xiaoyu Dong and Xuangeng Chu is with the University of Tokyo, Japan.}
\thanks{Lin Gu is with the Tohoku University, Japan.}
\thanks{Ming-Hsuan Yang is with the University of California at Merced and Google DeepMind, US.}
\thanks{Tatsuya Harada is with the University of Tokyo and RIKEN AIP, Japan.}

\thanks{\Envelope \ Corresponding Author: Lin Gu (lingu.edu@gmail.com)}}


\markboth{}%
{Shell \MakeLowercase{\textit{et al.}}: A Sample Article Using IEEEtran.cls for IEEE Journals}

\IEEEpubid{}

\maketitle

\begin{abstract}

High-quality image acquisition in real-world environments remains challenging due to complex illumination variations and inherent limitations of camera imaging pipelines. These issues are exacerbated in multi-view capture, where differences in lighting, sensor responses, and image signal processor (ISP) configurations introduce photometric and chromatic inconsistencies that violate the assumptions of photometric consistency underlying modern 3D novel view synthesis (NVS) methods, including Neural Radiance Fields (NeRF) and 3D Gaussian Splatting (3DGS), leading to degraded reconstruction and rendering quality.
We propose Luminance-GS++, a 3DGS-based framework for robust NVS under diverse illumination conditions. Our method combines a globally view-adaptive lightness adjustment with a local pixel-wise residual refinement for precise color correction. We further design unsupervised objectives that jointly enforce lightness correction and multi-view geometric and photometric consistency. Extensive experiments demonstrate state-of-the-art performance across challenging scenarios, including low-light, overexposure, and complex luminance and chromatic variations. Unlike prior approaches that modify the underlying representation, our method preserves the explicit 3DGS formulation, improving reconstruction fidelity while maintaining real-time rendering efficiency.

\end{abstract}


\section{Introduction}
\label{sec1:intro}

\IEEEPARstart{I}{n} recent years, Neural Radiance Fields (NeRF) and 3D Gaussian Splatting (3DGS)-based methods~\cite{nerf,NeRF++,Li_2023_ICCV,barron2023zipnerf,barron2022mipnerf360,3dgs,charatan23pixelsplat,Niedermayr_2024_CVPR,liu2025mvsgaussian} have substantially advanced realistic 3D scene reconstruction and photorealistic novel view synthesis. These approaches learn continuous scene representations from multi-view observations, enabling accurate geometry recovery and modeling of view-dependent appearance, and have become foundational techniques across applications such as AR/VR, computational photography, medical imaging, and autonomous driving.
However, existing methods rely heavily on implicit photometric consistency assumptions that rarely hold in real-world capture. Variations in illumination intensity (e.g., low-light and exposure differences), illumination color (e.g., color temperature shifts), and camera-dependent processing (e.g., white balance and ISP pipelines) introduce systematic photometric and chromatic inconsistencies, leading to degraded reconstruction fidelity and unstable rendering. These failures primarily arise from lightness distortions induced by environmental conditions and sensor limitations~\cite{raw_nerf,TOG_wang2024bilateral}, along with the limited capability of current NeRF- and 3DGS-based frameworks to model illumination variations explicitly~\cite{rudnev2022nerfosr,CVPR24_GS_IR}.

Numerous approaches have been proposed for lightness correction and perceptual image enhancement. Image enhancement methods~\cite{Zero-DCE,cvpr22_sci,ICCV2023_NeRCo,Wang_2024_CVPR,DiffRetiNex++} primarily address low-light conditions by increasing brightness while suppressing sensor noise. Exposure correction methods~\cite{Afifi_2021_CVPR,Cui_2022_BMVC,zhou2024mslt} aim to recover balanced luminance from underexposed or overexposed inputs, while white balance (WB) correction techniques compensate for illumination color shifts in either RAW~\cite{CCC_WB,Hu_FC_4} or sRGB domains~\cite{CNN_WB,WB_Flow,wb_color_space}.
Despite their success in single-image restoration, these methods are not well suited for multi-view 3D reconstruction. Prior work~\cite{LLNeRF,cui_aleth_nerf,zou2024enhancing_aaai} shows that independently preprocessing views disrupts radiometric consistency, violating key assumptions of NVS optimization. The resulting inter-view photometric discrepancies often manifest as floaters, ghosting artifacts, and geometric instability~\cite{SpotLessSplats_TOG}.


To improve NVS robustness under challenging lighting conditions, NeRF-W~\cite{martinbrualla2020nerfw} and subsequent works~\cite{HA-NeRF,yang2023CRNeRF,GS-W_ECCV2024,kulhanek2024wildgaussians} introduce view-dependent appearance embeddings to disentangle illumination effects from scene geometry. While effective at mitigating view inconsistencies and reducing floaters, these approaches are inherently constrained by the lighting conditions observed in the training views, limiting their ability to generalize to novel illumination scenarios during rendering (e.g., low-light enhancement or color correction).

\begin{figure*}[tp]
    \centering
    \includegraphics[width=1.0\linewidth]{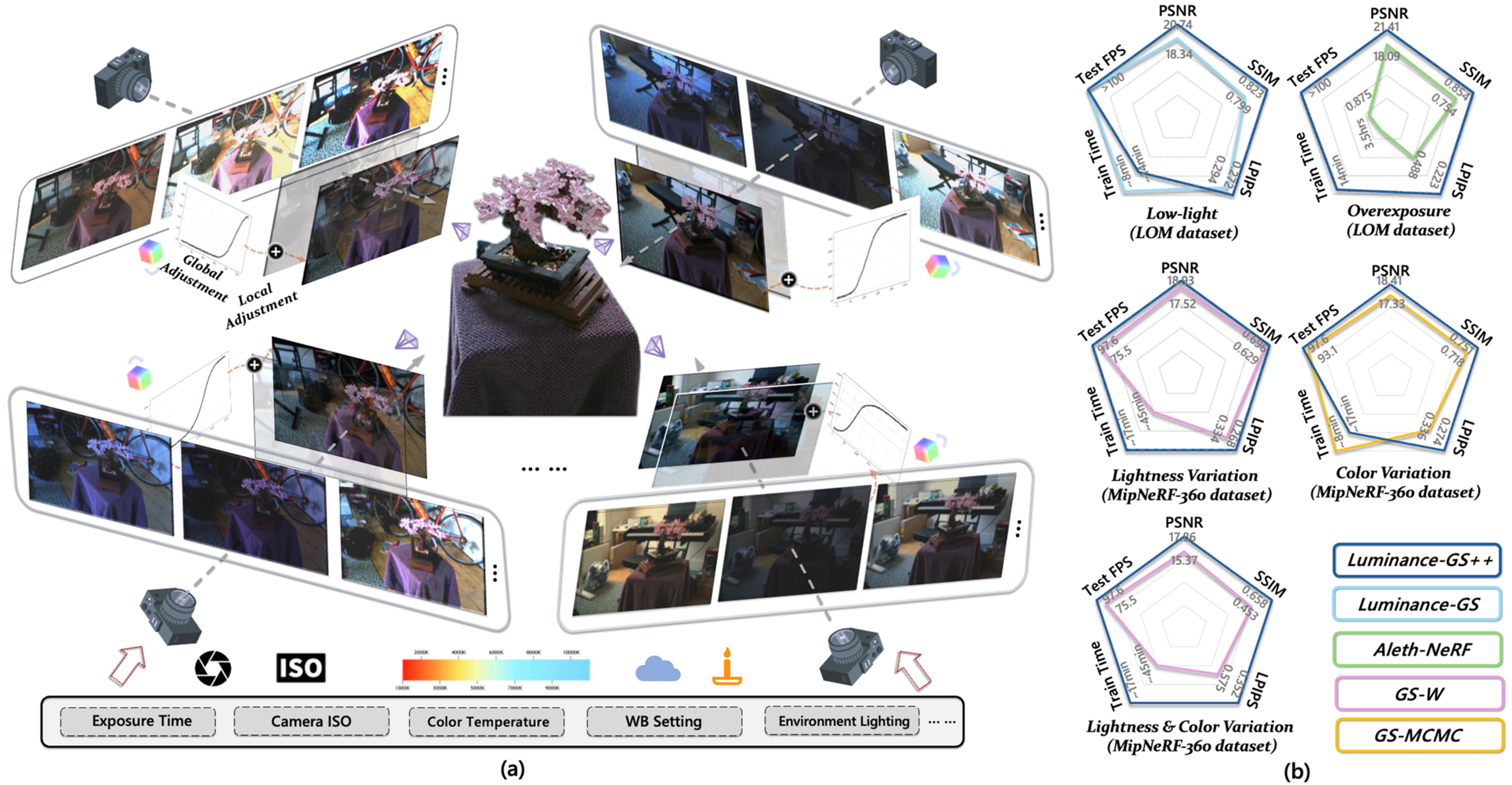}
    \vspace{-5mm}
    \caption{(a) An overview of \textbf{Luminance-GS++}, designed to address multi-view images affected by real-world illumination and color degradations via joint \textit{\textbf{global}} and \textit{\textbf{local}} adjustments, enabling robust 3DGS reconstruction. 
    (b) Performance comparison between \textbf{Luminance-GS++} and state-of-the-art methods~\cite{cui_luminance_gs,cui_aleth_nerf,GS-W_ECCV2024,MCMC} under diverse illumination and color conditions.}
    \label{fig:first}
    \vspace{-4mm}
\end{figure*}

To achieve robust lightness correction in multi-view 3D scenes, several NeRF-based approaches~\cite{LLNeRF,cui_aleth_nerf,zou2024enhancing_aaai,TOG_wang2024bilateral} incorporate illumination-specific modeling within the volume rendering pipeline, enabling unsupervised lightness adjustment while preserving multi-view consistency. LL-NeRF~\cite{LLNeRF} decomposes the NeRF color MLP into view-dependent and view-independent components for targeted processing. Aleth-NeRF~\cite{cui_aleth_nerf} introduces a concealing-field branch that models low-light effects as occlusions, allowing adaptive enhancement or suppression during inference. AME-NeRF~\cite{zou2024enhancing_aaai} formulates a bilevel optimization framework to normalize brightness while maintaining structural consistency. For multi-view color correction, Zhang \textit{et al.}~\cite{NeRF_CC_ECCV24} emulate digital camera processing by integrating ISP-inspired kernel mappings and color matrices into NeRF’s implicit representation.

However, these solutions do not readily extend to explicit representations such as 3DGS~\cite{3dgs}, which relies on rasterization-based rendering and explicitly stores appearance within a 3D Gaussian point cloud, eliminating the need for implicit neural fields and MLP-based architectures. This explicit formulation provides substantial advantages over NeRF-based methods in both training and rendering efficiency~\cite{3dgs,charatan23pixelsplat,HDR_GS_nips,fan2023lightgaussian}. For example, compared to state-of-the-art NeRF approaches such as~\cite{barron2022mipnerf360}, 3DGS achieves over $1000\times$ faster rendering and nearly $480\times$ faster training while maintaining comparable visual quality~\cite{3dgs}. These advances raise a fundamental question: how can we \textit{leverage the efficiency of explicit 3DGS representations} while \textit{ensuring robust reconstruction} under diverse and challenging illumination conditions?

Our preliminary work, Luminance-GS~\cite{cui_luminance_gs}, introduces a pseudo-label–based image enhancement strategy that jointly optimizes tonal curves with 3D Gaussian attributes. Specifically, the method applies \textit{{per-view color matrix mapping}} and \textit{{view-adaptive curve adjustments}} via differentiable tone curves to transform inputs captured under diverse lighting conditions (e.g., low-light, overexposure, or exposure variations) into consistently well-lit pseudo-enhanced images. These color transformations are jointly optimized with the Gaussian representation during training, while unsupervised objectives enforce multi-view alignment and radiometric consistency across viewpoints.



However, Luminance-GS assumes white ambient illumination and does not explicitly evaluate reconstruction robustness under varying color temperatures or mixed degradations. Moreover, it relies on a \textbf{{global}} curve-adjustment strategy for pseudo-label generation, applying view-level tone operations that lack \textbf{{local}} pixel-wise refinement capabilities (e.g., pixel-level white balance correction or handling specular residuals).

To address these limitations, we propose the upgraded {Luminance-GS++}, which introduces an additional residual branch for pixel-level local adjustments, improving decomposition fidelity (see Fig.~\ref{fig:first}(a)). We further redesign the loss functions to strengthen unsupervised multi-view color correction and conduct extensive new experiments to validate the effectiveness of the proposed approach. The main contributions of our work are summarized as follows:
\begin{itemize}
    \item We propose {Luminance-GS++}, a framework that improves the robustness of novel view synthesis (NVS) under lightness and color degradations. Without modifying the explicit 3DGS representation, our method introduces a pseudo-label generation strategy that integrates \textit{{global adjustment}} and \textit{{local refinement}}, jointly optimized with the 3DGS model.

    \item We design unsupervised objectives for multi-view lightness and color correction that preserve radiometric consistency, enabling {Luminance-GS++} to generalize across diverse degradation scenarios while maintaining stable 3D reconstruction.

    \item We perform extensive evaluations on benchmark datasets under challenging real-world illumination and synthetic lighting variations. Extensive eperimental results demonstrate consistent improvements over state-of-the-art 3D NVS methods.
\end{itemize}



     
     

\section{Related Work}
\label{sec2:related}




\subsection{3D Gaussian Splatting and Its Applications}

In recent years, Neural Radiance Fields (NeRF)~\cite{nerf} have significantly advanced novel view synthesis (NVS). However, NeRF-based methods~\cite{nerf,Li_2023_ICCV,barron2023zipnerf,barron2022mipnerf360,chugunov2024light,NEURIPS2022_fe989bb0} are computationally expensive, as the  rendering requires dense sampling and multiple network evaluations per pixel, limiting real-time applicability. 
Most recently, 3D Gaussian Splatting (3DGS)~\cite{3dgs} has emerged as a highly efficient alternative, enabling real-time NVS while maintaining rendering quality comparable to state-of-the-art NeRF-based approaches~\cite{barron2022mipnerf360}. Its efficiency arises from an explicit scene representation based on learnable anisotropic 3D Gaussians and differentiable splatting via tile-based rasterization.

Building on this foundation, numerous works have extended 3DGS to improve rendering quality~\cite{guedon2023sugar,Huang2DGS2024,Fu_2024_CVPR,MCMC}, training efficiency~\cite{fan2023lightgaussian,iclr25_scale_up_gs,3dgs_prune_nips}, generalization capability~\cite{charatan23pixelsplat,CVPR24_splatter_image,chen2024mvsplat,liu2025mvsgaussian}, and real-world applicability~\cite{CVPR24_GS_IR,X_Ray_GS_ECCV,li2024sgs,zhou2024drivinggaussian,lidense2025,liu2025mg,Gaussian_Wave}. Representative examples include 2DGS~\cite{Huang2DGS2024}, which replaces volumetric Gaussians with oriented 2D disks to better capture surface structure; LightGaussian~\cite{fan2023lightgaussian}, which introduces a pruning–recovery pipeline for efficient compression; GS-MCMC~\cite{MCMC}, which models Gaussians as samples from a scene distribution optimized via stochastic gradient Langevin dynamics; and MVSplat~\cite{chen2024mvsplat}, which leverages Transformer-based multi-view features with cost-volume construction for feed-forward reconstruction. Application-driven extensions further demonstrate the versatility of 3DGS, including X-Gaussian~\cite{X_Ray_GS_ECCV} for sparse-view CT reconstruction, DrivingGaussian~\cite{zhou2024drivinggaussian} for dynamic driving scenes, and Gaussian Wave Splatting~\cite{Gaussian_Wave} for occlusion-aware holographic rendering.

\subsection{3D Reconstruction under Degradation} 

\subsubsection{Robust 3D Reconstruction} 
Despite significant progress in NeRF- and 3DGS-based reconstruction, most existing methods assume ideal imaging conditions. In real-world scenarios involving degradations such as low resolution~\cite{wang2021nerf-sr,wan2025s2gaussian}, underwater environments~\cite{watersplatting,seathru}, fog~\cite{scatternerf,liu2025_i2nerf}, rain~\cite{Derain_GS}, noise~\cite{NAN_CVPR_2022}, and motion blur~\cite{deblur_nerf,deblur_3dgs}, reconstruction performance often degrades significantly~\cite{liu2025_realx3d,lin2025hqgs}. In this work, we focus on degradations induced by illumination and color variations (e.g., environmental lighting changes, ISO settings, exposure variations, color temperature shifts, and white balance errors), which are pervasive yet relatively underexplored challenges in real-world multi-view capture.

\subsubsection{Lightness Degradations} 
For real-world lightness degradations, methods such as NeRF-W~\cite{martinbrualla2020nerfw} and its follow-up works~\cite{HA-NeRF,yang2023CRNeRF,GS-W_ECCV2024,Dahmani2024SWAGSI,kulhanek2024wildgaussians} address reconstruction under inconsistent illumination and occlusions by disentangling appearance variations from scene geometry. This line of research primarily aims to recover uniform lighting from multi-view inputs exhibiting non-uniform illumination. For example, NeRF-W~\cite{martinbrualla2020nerfw} introduces learnable appearance and transient embeddings per view, while GS-W~\cite{GS-W_ECCV2024} models intrinsic and dynamic appearance components for individual 3D Gaussian primitives. However, these approaches are largely limited to reconstructing lighting conditions observed in the training data, restricting adaptation to novel illumination scenarios (e.g., low-light enhancement).

For low-light environments, RAW-NeRF~\cite{raw_nerf} leverages HDR RAW inputs combined with an ISP model for improved reconstruction. Subsequent works~\cite{zhang2024darkgs,jin2024le3d,li2024chaosclarity3dgsdark,singh24_hdrsplat} extend this paradigm to 3DGS for improved efficiency. Nevertheless, reliance on RAW data introduces practical limitations, including acquisition constraints, storage overhead, and prolonged training times.

Several studies have instead explored NVS under challenging lighting using sRGB inputs~\cite{LLNeRF,cui_aleth_nerf,xu2024leveraging,zou2024enhancing_aaai,ye2024gaussian_in_dark,TOG_wang2024bilateral,qu2024lushnerf,lita-gs,liu2025_i2nerf}. For instance, Aleth-NeRF~\cite{cui_aleth_nerf} introduces a concealing field to regulate image lightness, enabling both low-light enhancement and overexposure suppression. BilaRF~\cite{TOG_wang2024bilateral} employs 3D bilateral grids to model view-specific camera effects for refined tonal adjustment. Thermal-NeRF~\cite{xu2024leveraging} fuses thermal and sRGB data for improved reconstruction under low-light conditions. In contrast, our Luminance-GS++ adopts a curve-based pseudo-enhancement framework that generalizes across diverse lighting conditions without requiring modifications to training strategies or hyperparameters.

\subsubsection{Color Distortions} 
Real-world color distortions caused by varying color temperatures and camera white balance errors remain relatively underexplored. Early works~\cite{Multi_view_CC_1,Multi_view_CC_2,Multi_view_CC_3,Gu_TIP12,Gu_TIP14} focused primarily on multi-camera color consistency rather than addressing challenges in 3D NVS. URF~\cite{rematas2022urf} applies a per-image latent-decoded $3\times3$ matrix for white balance correction, but its global formulation limits flexibility under complex color variations. Zhang \textit{et al.}~\cite{NeRF_CC_ECCV24} propose a physically grounded NeRF-based color correction framework incorporating nonlinear modules and histogram-based consistency losses; however, the histogram loss does not explicitly enforce 3D consistency, and reliance on implicit NeRF representations results in slower training. Our approach instead leverages the efficiency of explicit 3DGS~\cite{3dgs} to achieve robust and efficient reconstruction under color distortions.

\subsection{Curve Adjustment in Image Processing}
Curve-based adjustment modifies tonal ranges by mapping pixel intensities through predefined or learned tone curves, commonly used in commercial tools such as Adobe Lightroom$^\circledR$. Early image enhancement methods relied on handcrafted tone-mapping functions (e.g., power-law curves and S-curves) that required manual tuning to achieve desirable perceptual results~\cite{Siggraph2002_curve,curve_cvpr2010,Curve_old,CVPR11_curve}.

More recently, data-driven approaches learn tone curves from large-scale datasets for global or local image enhancement~\cite{Zero-DCE,SACC_2022_ACMMM,jiang2023meflut,moran2020curl,Curve_prediction_ECCV,curve_TOG,yang2023difflle,eccv24_colorcurves,lee2024cliptone,cuibmvc2024}. For example, CURL~\cite{moran2020curl} learns tone adjustments across multiple color spaces, while Zero-DCE~\cite{Zero-DCE} predicts pixel-wise high-order curves using unsupervised objectives. NamedCurves~\cite{eccv24_colorcurves} further decomposes images via color naming systems and applies Bézier curves per category. Unlike these 2D methods, our approach performs view-dependent curve optimization across multi-view inputs, enabling consistent brightness and color correction while preserving 3D consistency for stable 3DGS training.

\begin{figure}[tp]
    \centering
    \includegraphics[width=1.0\linewidth]{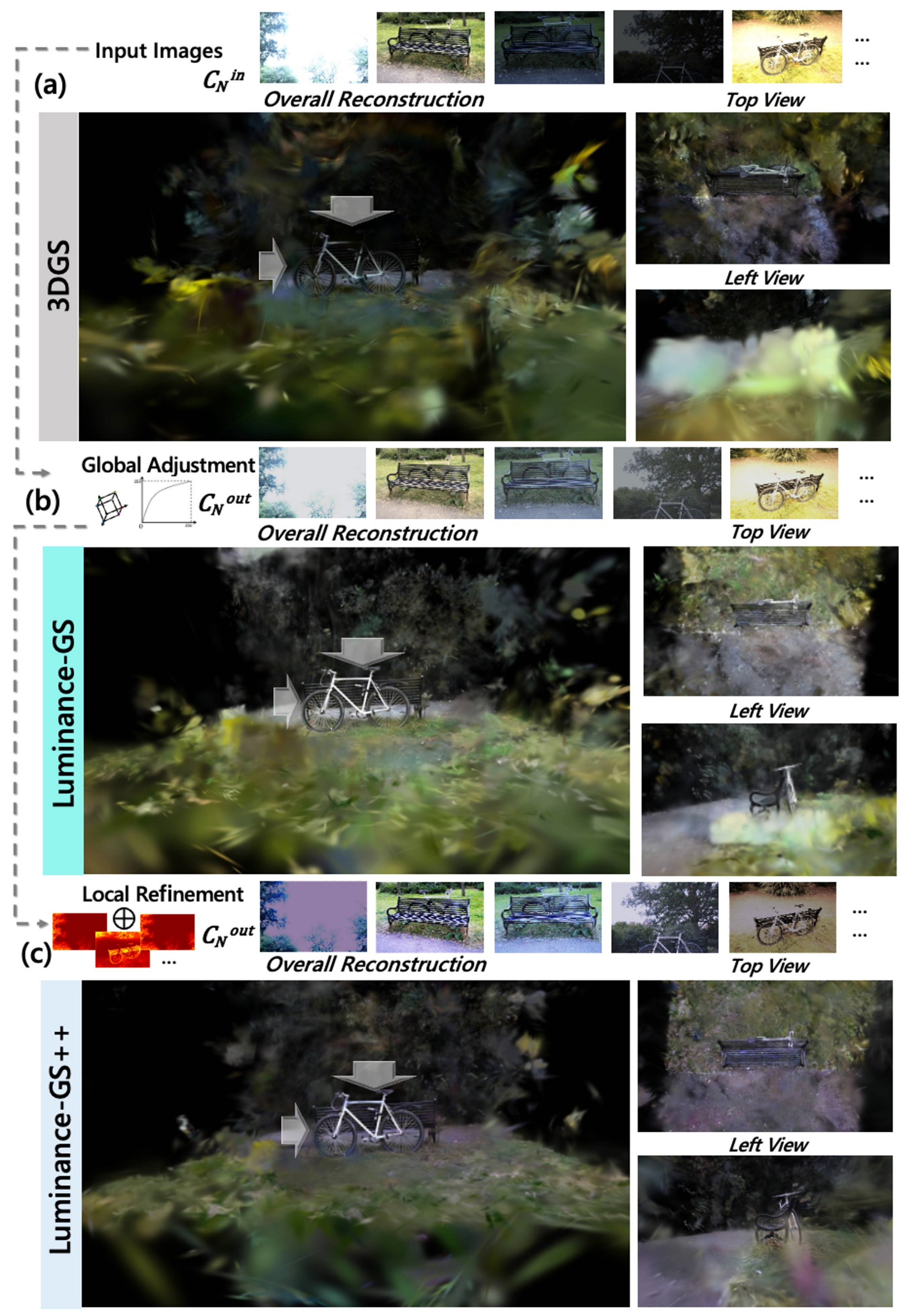}
    \caption{Progression from 3DGS~\cite{3dgs} to Luminance-GS~\cite{cui_luminance_gs} and Luminance-GS++, highlighting joint optimization of input images and 3DGS attributes through global adjustment and local refinement. Multi-view rendering examples illustrate the resulting improvements.}
    \label{fig:ablation}
    \vspace{-2mm}
\end{figure}

\section{Proposed Method}
\label{sec3:method}

We begin with a brief overview of the vanilla 3D Gaussian Splatting (3DGS) framework~\cite{3dgs} (Sec.~\ref{sec3.1}). We then describe the key training enhancements that led to {Luminance-GS++} (Sec.~\ref{sec3.2}), followed by a detailed presentation of the proposed architecture and pipeline, which integrate {\textit{global}} curve adjustment with {\textit{local}} pixel-wise refinement (Sec.~\ref{sec3.3}).

\subsection{3D Gaussian Splatting Revisited}
\label{sec3.1}

3DGS~\cite{3dgs} is an explicit primitive-based representation that models a scene as a collection of anisotropic 3D Gaussians $\left\{ G_1, ..., G_M \right\}$. Each Gaussian $G_i$ is parameterized by a center position $\mu_i \in \mathbb{R}^3$, a covariance matrix $\Sigma_i \in \mathbb{R}^{3 \times 3}$, view-dependent color $c_i$, and opacity $o_i \in [0,1]$. The covariance matrix encodes spatial extent and orientation and is decomposed into a diagonal scaling matrix $S_i = \mathrm{diag}(\sigma_{i1}, \sigma_{i2}, \sigma_{i3})$ and a rotation matrix $R_i \in SO(3)$:
\begin{equation}
\Sigma_i = R_i S_i^2 R_i^\top.
\end{equation}
This formulation represents each Gaussian as a fully anisotropic ellipsoid in 3D space. View-dependent appearance is modeled using low-order spherical harmonics, enabling efficient approximation of angular radiance variation and supporting real-time rendering.

During rendering, Gaussians are projected onto the image plane using camera parameters and accumulated via differentiable elliptical surface splatting~\cite{zwicker2001surface}. Projection produces a 2D elliptical footprint determined by the transformed covariance $\Sigma_i^{\text{screen}} = J_i \Sigma_i J_i^\top$, where $J_i$ denotes the projection Jacobian evaluated at $\mu_i$. Each pixel $x$ receives contributions from overlapping Gaussians, sorted by increasing depth. The rendered color is computed using front-to-back compositing:
\begin{equation}
    \hat{C}(x) = \sum_{i=1}^N c_i o_i' \prod_{j=1}^{i-1} (1 - o_j'),
    \label{Eq:rendering}
\end{equation}
where $o_i'$ denotes the visibility-weighted opacity modulated by the projected Gaussian footprint and pixel distance. This formulation resembles volumetric alpha compositing and enables smooth blending and partial transparency~\cite{nerf}.

\begin{figure}
    \centering
    \includegraphics[width=1.00\linewidth]{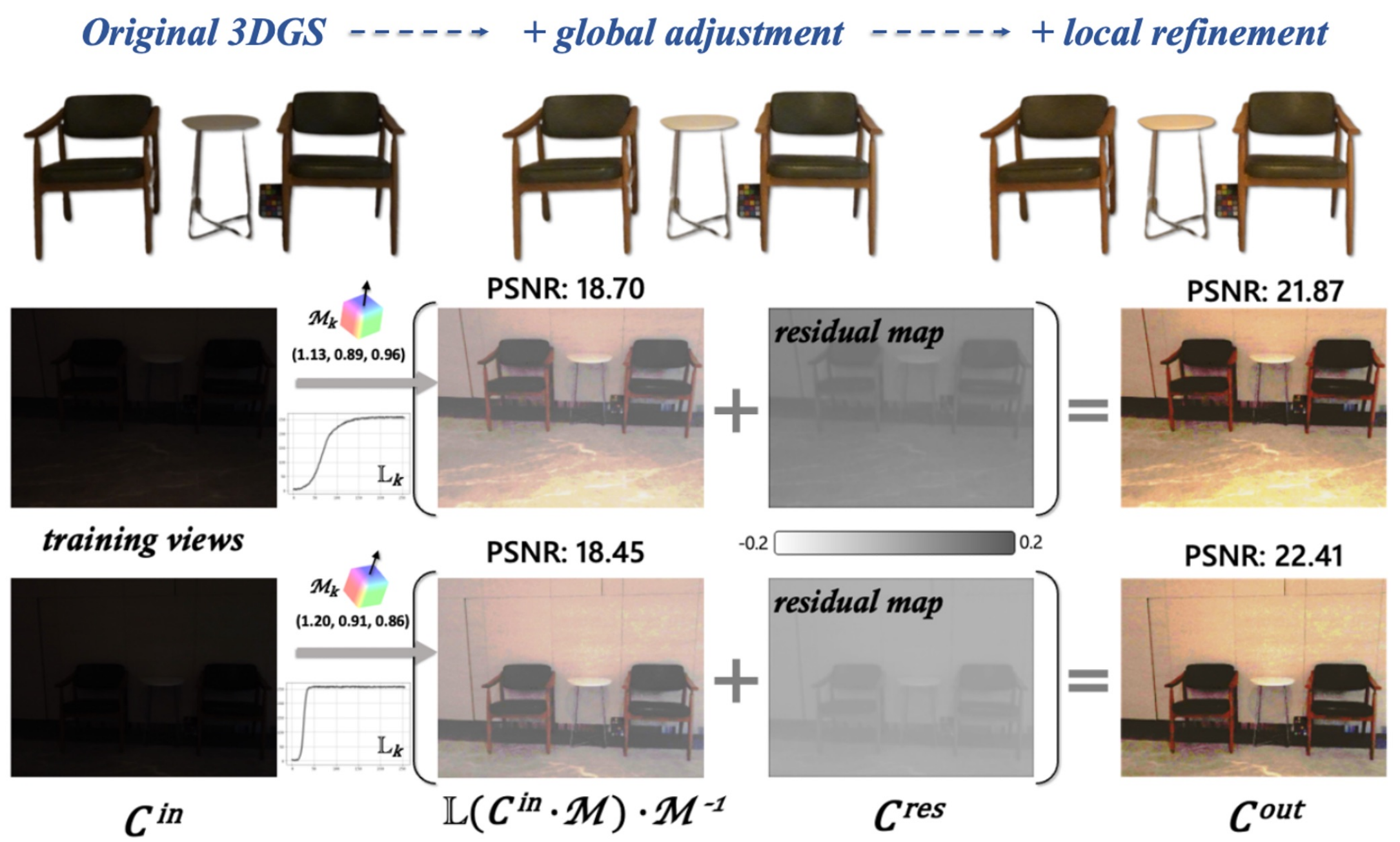}
    \vspace{-3mm}
    \caption{A low-light \textit{\textbf{``chair''}} scene from the LOM dataset~\cite{cui_aleth_nerf}. Our global adjustment enhances overall illumination, while our local refinement improves fine details and the quality of pseudo label $C^{out}$.}
    \label{fig:ablation_lom}
    \vspace{-2mm}
\end{figure}

All Gaussian parameters are optimized using a mixed reconstruction objective $\mathcal{L}_{\rm 3DGS}$ that measures the discrepancy between the rendered image $\hat{C}$ and ground truth image $C$:
\begin{equation}
    \mathcal{L}_{\rm 3DGS}(\hat{C}, C) = \lambda \mathcal{L}_{\rm DSSIM}(\hat{C}, C) + (1 - \lambda) \mathcal{L}_1(\hat{C}, C),
    \label{Eq:3d_gs_loss}
\end{equation}
where $\mathcal{L}_{\rm DSSIM}$ denotes DSSIM loss, $\mathcal{L}_1$ is pixel-wise L1 loss, and $\lambda \in [0,1]$ balances the two terms. The differentiable rendering pipeline enables joint optimization of all Gaussian parameters for high-fidelity multi-view reconstruction.

Despite its efficiency, 3DGS performs direct color fitting, where spherical harmonics provide only an approximate shading representation without explicit decomposition of reflectance and illumination~\cite{jin2024le3d,CVPR24_GS_IR}. Consequently, under illumination degradations, the model entangles material and lighting effects, producing floaters and color shifts in rendered outputs (see Fig.~\ref{fig:ablation}(a)).

\subsection{From 3DGS to Luminance-GS++}
\label{sec3.2}
  
\begin{figure*}[tp]
    \centering
    \includegraphics[width=0.95\linewidth]{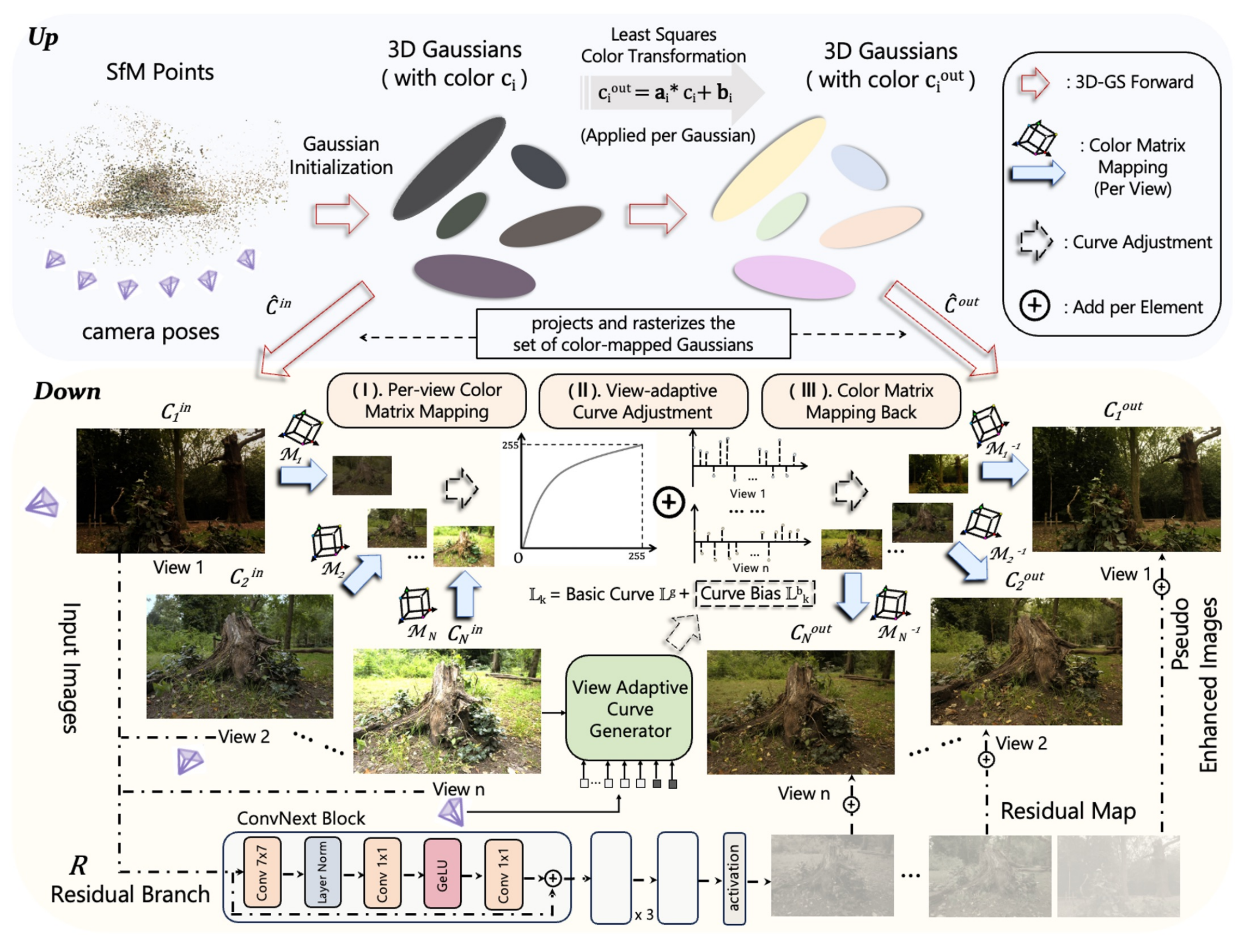}
    \vspace{-4mm}
    \caption{\textbf{An overview of the Luminance-GS++ pipeline}. 
    \textbf{\textit{Up}}: Joint optimization of 3D Gaussians with dual color attributes $c_i$ and $c_i^{out}$ to render input images $C^{in}$ and pseudo-enhanced outputs $C^{out}$. 
    \textbf{\textit{Down}}: Pseudo-enhancement is achieved through \textit{global adjustment} (view-adaptive color matrix mapping and curve adjustment; stages (\uppercase\expandafter{\romannumeral1})--(\uppercase\expandafter{\romannumeral3})) followed by \textit{local refinement} using a pixel-wise residual map.
    }
    \label{fig:overview}
    \vspace{-3mm}
\end{figure*}

Figure~\ref{fig:overview} provides an overview of the proposed {Luminance-GS++} framework. Starting from SfM points~\cite{COLMAP2} (Fig.~\ref{fig:overview} \textbf{\textit{Up}}), we initialize a set of anisotropic 3D Gaussians $\{ G_1, \dots, G_M \}$ with color attributes $c_i$ to reconstruct multi-view images $\{ C^{in}_1, \dots, C^{in}_N \}$ captured under challenging illumination conditions. In addition to the standard 3DGS representation, we introduce learnable color adjustment parameters $\mathbf{a}_i$ and $\mathbf{b}_i$, which transform the original color attributes into adjusted colors using a linear mapping~\cite{kulhanek2024wildgaussians}:
\begin{equation}
    c_i^{out} = \mathbf{a}_i \cdot c_i + \mathbf{b}_i.
\end{equation}

The transformed colors $c_i^{out}$ are used to render pseudo-enhanced images $C^{out}$ alongside the original colors $c_i$. During training, Luminance-GS++ jointly predicts both the original and pseudo-enhanced renderings, extending Eq.~\ref{Eq:rendering} as:
\begin{equation}
\begin{aligned}
    \hat{C}^{in}(x) &= \sum_{i=1}^N c_i o_i' \prod_{j=1}^{i-1} (1 - o_j'), \\
    \hat{C}^{out}(x) &= \sum_{i=1}^N c_i^{out} o_i' \prod_{j=1}^{i-1} (1 - o_j').
\label{Eq:rendering_ours}
\end{aligned}
\end{equation}
During inference, the rendering relies solely on $c_i^{out}$, preserving the real-time performance of the original 3DGS pipeline.

The pseudo-enhanced image generation process (Fig.~\ref{fig:overview} \textbf{\textit{Down}}) aims to address real-world degradations including illumination variations (e.g., low-light, overexposure) and chromatic distortions (e.g., varying color temperatures). Since 3DGS training implicitly assumes consistent lighting, our objectives are twofold: (1) normalize illumination toward well-exposed conditions, and (2) enforce cross-view consistency to stabilize multi-view optimization.

Figure~\ref{fig:ablation}(a) illustrates direct 3DGS reconstruction on a scene~\cite{barron2022mipnerf360}, where illumination and color degradations introduce view-inconsistent appearances, 
preventing reliable surface alignment~\cite{kulhanek2024wildgaussians,SpotLessSplats_TOG}. As a result, Gaussians may stretch or drift in space, producing elongated or floating artifacts.

Our preliminary work, Luminance-GS~\cite{cui_luminance_gs}, applies global tonal curve adjustment by mapping the color matrix per-view and view-adaptive curves (Fig.~\ref{fig:ablation}(b)). While effective for normalizing exposure and illumination, global curve operations alone cannot fully resolve multi-view color inconsistencies, leading to residual artifacts. In addition, curve-based mappings apply identical transformations to pixels with identical values~\cite{Curve_prediction_ECCV}, limiting spatial adaptivity.

To overcome these limitations, {Luminance-GS++} introduces local pixel-wise refinement following global adjustment. A learned residual map enables fine-grained correction of spatially varying color and illumination discrepancies, improving pseudo-enhanced image quality and reconstruction stability (Fig.~\ref{fig:ablation}(c)). Even under relatively uniform lighting conditions (e.g., low-light scenes), local refinement further enhances detail fidelity, as demonstrated in Fig.~\ref{fig:ablation_lom}.

\subsection{Module Details and Pipeline}
\label{sec3.3}

This section describes the generation of pseudo-enhanced images $C^{out}$ from multi-view inputs $C^{in}$ and details the proposed \textit{global adjustment} and \textit{local refinement} modules.

\subsubsection{Global Adjustment}

The \textit{global adjustment} module follows the design in Luminance-GS~\cite{cui_luminance_gs}, consisting of per-view color matrix mapping and view-adaptive curve adjustments.


Specifically,
to compensate for multi-view lightness variations caused by illumination differences and camera exposure settings~\cite{GS-W_ECCV2024,kulhanek2024wildgaussians}, we introduce a learnable per-view color transformation matrix $\mathcal{M}_k$. Before curve adjustment, each input image $C^{in}_{k \in (1,N)}$ is projected into a view-dependent color space via $\mathcal{M}_k$. Figure~\ref{fig:detail_block}(a) shows an illustration of this color matrix mapping.

As indicated by the blue arrow in Fig.~\ref{fig:overview}, given an input image $C^{in}_k(r,g,b)$ in RGB color space $\mathbb{I}(R,G,B)$, we optimize a per-view invertible $3\times3$ matrix $\mathcal{M}_k$, initialized as the identity matrix and jointly optimized with the 3DGS parameters. The mapping is defined as:
\begin{equation}
\begin{aligned}
  C^{in}_k \cdot \mathcal{M}_k
  &= \left[C^{in}_k(r), C^{in}_k(g), C^{in}_k(b) \right]
  \begin{bmatrix}
   a_{11} & a_{12} & a_{13}  \\
   a_{21} & a_{22} & a_{23} \\
   a_{31} & a_{32} & a_{33}
  \end{bmatrix}, \\
  &= \left[C^{in}_k(r'), C^{in}_k(g'), C^{in}_k(b') \right],
\end{aligned}
\label{eq:matrix_trans}
\end{equation}
where each element $\mathcal{M}_k(a_{ij})$ is learnable. The transformed image $C^{in}_k(r',g',b')$ represents the projection into a view-dependent coordinate system for subsequent tone adjustment.

For tone mapping, we learn a shared global curve ${\mathbb L}^g$ and a per-view curve bias ${\mathbb L}_k^b$, yielding the final tone curve ${\mathbb L}_k = {\mathbb L}^g + {\mathbb L}_k^b$. The global curve enforces consistent brightness and color tone across views, while the view-specific bias provides flexibility under non-uniform illumination.

In practice, ${\mathbb L}^g$ is implemented as a 1D lookup table of length 256. To prevent overfitting, ${\mathbb L}_k^b$ is generated via a view-adaptive attention module inspired by~\cite{DETR,Cui_2022_BMVC}. As illustrated in Fig.~\ref{fig:detail_block}(b), the image $C^{in}_k$ is encoded using convolutional and linear layers to produce key and value representations, while the camera matrix serves as the query. Cross-attention~\cite{dosovitskiy2021an} followed by a feed-forward network outputs a $1\times256$ curve bias.

The final tone curve ${\mathbb L}_k$ is applied to the mapped image $C^{in}_k \cdot \mathcal{M}_k$, and the result is transformed back to RGB space using $\mathcal{M}_k^{-1}$, producing the pseudo-enhanced image:
\[
C^{out}_k = {\mathbb L}_k(C^{in}_k \cdot \mathcal{M}_k)\cdot \mathcal{M}_k^{-1}.
\]
Despite its effectiveness, global curve adjustment alone cannot fully resolve viewpoint-dependent color variations (Fig.~\ref{fig:ablation}(b)), and lacks pixel-level adaptability for fine detail correction (Fig.~\ref{fig:ablation_lom}). To address these limitations, we introduce a local refinement branch described next.


\subsubsection{Local Refinement}
The bottom part of Fig.~\ref{fig:overview} illustrates the structure of the \textit{{local refinement}} residual branch ${\mathbb R}$. By adding an additional pixel-wise residual map $C^{res}_{k}$ to the enhanced pseudo-labeled images, we aim to improve the model’s robustness to color variations while further refining fine-grained details. For each input view $C^{in}_{k}$, the image is passed through the residual branch to generate its corresponding residual map $C^{res}_{k}$.

The residual branch ${\mathbb R}$ consists of three ConvNeXt blocks~\cite{liu2022convnet} connected in series, followed by an activation layer. Each ConvNeXt block consists of a 7×7 convolution followed by layer normalization (LN), a 1×1 convolution, a GeLU activation, and another 1×1 convolution. The output is subsequently fused with the block input through a residual connection. When each view input image $C^{in}_{k}$ is processed by the three ConvNeXt blocks, it is subsequently passed through an activation layer to obtain the final residual map $C^{res}_{k}$. The purpose of the final activation layer is to constrain the output values of the residual map. Specifically, we employ a clipping operation, where the range is set to $[-0.1,0.1]$ for lightness variation scenarios and 
$[-0.5,0.5]$ for color variation scenarios. 
The final pseudo-enhanced image $C^{out}_k$ is processed through \textit{{global adjustment}} and \textit{{local refinement}}, as expressed below:
\begin{equation}
    C^{out}_k = {\mathbb L}_k(C^{in}_k \cdot \mathcal{M}_k) \cdot \mathcal{M}_k^{-1} + {\mathbb R}(C^{in}_k).
\end{equation}
In the next section, we introduce the useful loss constraints that guide the pseudo-enhanced labels $C^{out}_k$ and the overall {Luminance-GS++} training.

\begin{figure}
    \centering
    \includegraphics[width=0.98\linewidth]{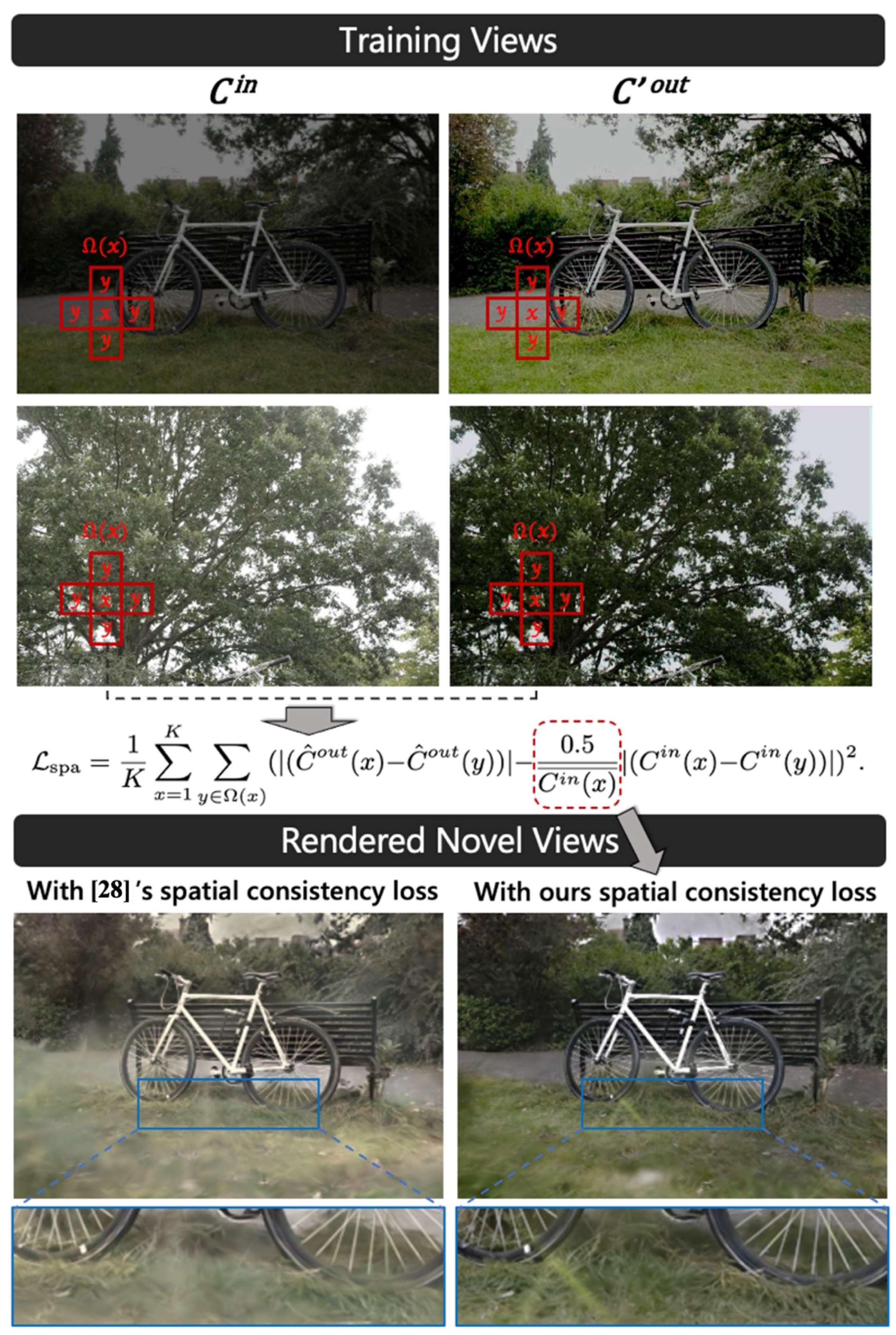}
    \vspace{-2mm}
    \caption{An illustration of our spatial consistency loss function $\mathcal{L}_{\rm spa}$, and comparisons with the spatial consistency loss in Aleth-NeRF~\cite{cui_aleth_nerf}.}
    \label{fig:Spa_Loss}
    \vspace{-4mm}
\end{figure}

\section{Loss Functions}
\label{sec4:augmentation}

The overall objective consists of three components: image intensity losses, image color loss, and curve regularization terms. The image-level losses supervise pseudo-label generation and the optimization of 3DGS-related attributes, while the curve-level constraints preserve the structural properties of the tone curve $\mathbb L$.

\begin{figure*}[tp]
    \centering
    \includegraphics[width=0.88\linewidth]{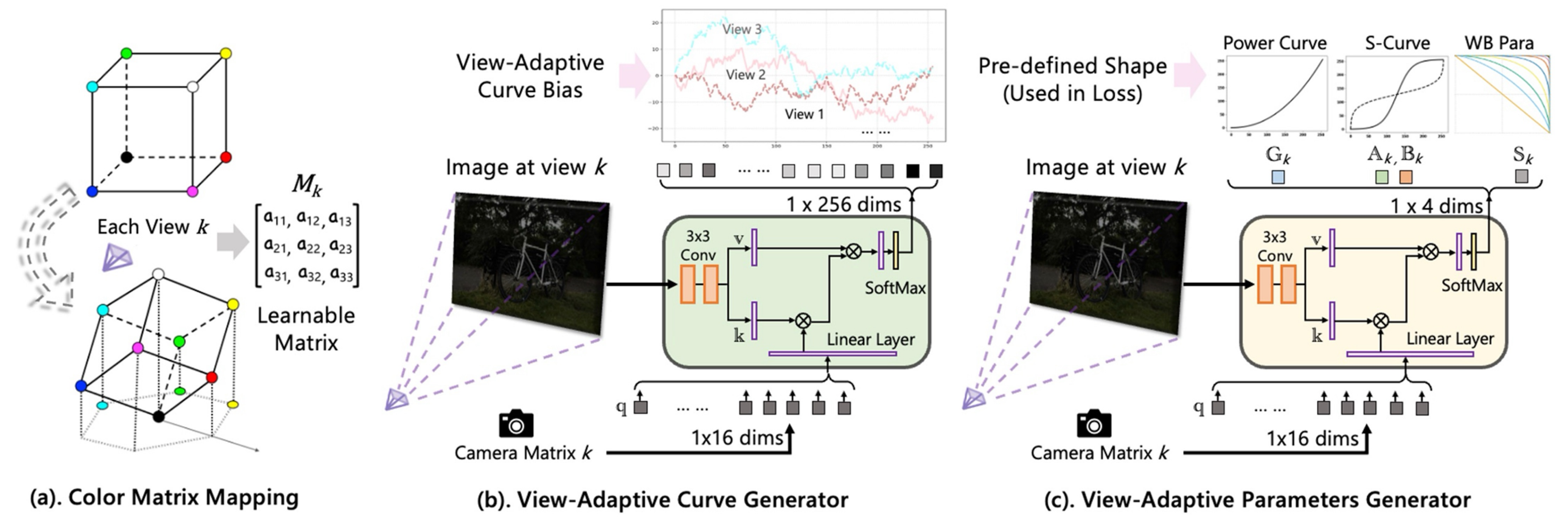}
    \caption{(a) Per-view color matrix mapping with learnable transformation $\mathcal{M}_k$. 
    (b) Architecture of the view-adaptive curve generator, which predicts curve bias ${\mathbb L}_k^b$ from input image $C^{in}_k$ and camera pose. 
    (c) Structure of the view-adaptive parameter generator.}
    \label{fig:detail_block}
    \vspace{-3mm}
\end{figure*}

\subsection{Image Intensity Losses}

We introduce the image-level objectives based on the dual-rendering formulation in Eq.~\ref{Eq:rendering_ours}. {Luminance-GS++} employs two sets of color attributes, $c_i$ and $c_i^{out}$, to jointly reconstruct the input multi-view images $C^{in}_{k \in (1,N)}$ and the pseudo-enhanced images $C^{out}_{k \in (1,N)}$. Accordingly, the overall regression loss extends Eq.~\ref{Eq:3d_gs_loss} as:
\begin{equation}
    \mathcal{L}_{\rm reg} = \mathcal{L}_{\rm 3DGS}(\hat{C}^{in}, C^{in}) + \mathcal{L}_{\rm 3DGS}(\hat{C}^{out}, C^{out}).
\end{equation}

To preserve structural consistency during enhancement, we further introduce a spatial consistency loss $\mathcal{L}_{\rm spa}$ inspired by Zero-DCE~\cite{Zero-DCE} and Aleth-NeRF~\cite{cui_aleth_nerf}. This term enforces consistency of local intensity relationships between the pseudo-enhanced image $\hat{C}^{out}$ and the input image $C^{in}$ by matching relative differences between neighboring pixels:
\begin{equation}
\begin{aligned}
    \mathcal{L}_{\rm spa} = & \frac{1}{K}\sum_{x=1}^{K}\sum_{y\in\Omega(x)}
    \Big(|\hat{C}^{out}(x)-\hat{C}^{out}(y)| \\
    & - \frac{0.5}{\overline{C^{in}(x)}} |C^{in}(x)-C^{in}(y)|\Big)^2,
\end{aligned}
\end{equation}
where $x$ denotes a pixel location and $y \in \Omega(x)$ represents its neighboring pixels. 
Unlike the fixed weighting in Aleth-NeRF~\cite{cui_aleth_nerf}, which can result in degraded novel view rendering quality (Fig.~\ref{fig:ablation} (bottom)), we introduce an adaptive factor $\frac{0.5}{\overline{C^{in}(x)}}$, where $\overline{C^{in}(x)}$ denotes the mean intensity of the current input view. This adaptive weighting improves robustness under varying illumination conditions. 

\begin{figure}[t]
    \centering
    \includegraphics[width=0.92\linewidth]{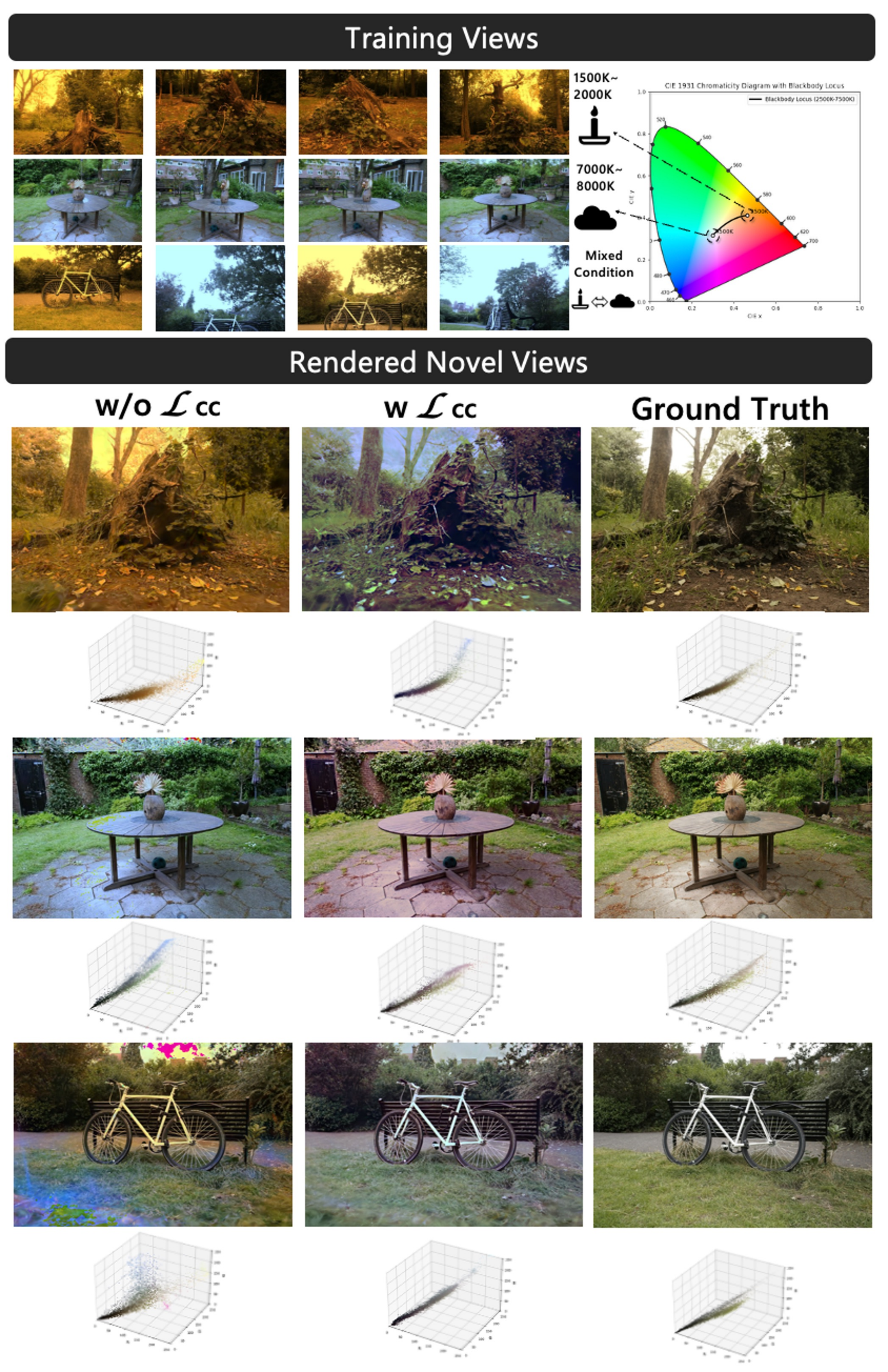}
    \caption{Effect of varying color temperatures (cool, warm, and mixed) on NVS. We compare the results with and without the color correction loss $\mathcal{L}_{\rm cc}$ and visualize the corresponding color distributions of rendered novel views using scatter plots.}
    \vspace{-3mm}
    \label{fig:color_loss}
    \vspace{-2mm}
\end{figure}

\subsection{Image Color Loss}

Multi-view color variation introduces additional challenges, as color temperature and camera white balance (WB) settings may either remain consistent across views or vary significantly (see Fig.~\ref{fig:color_loss}, \textit{{Up}}). Traditional single-view WB approaches rely on heuristic priors (e.g., gray-world~\cite{gray_world}, white-patch~\cite{white_patch}) or end-to-end learning strategies~\cite{WB_Flow,CNN_WB}, but typically operate on individual images without enforcing cross-view consistency.

Inspired by the Shade-of-Gray (SoG) algorithm, which generalizes classical WB methods through a tunable Minkowski parameter $\mathbb{S}$, we introduce a color correction loss $\mathcal{L}_{\rm cc}$:
\begin{equation}
    \mathcal{L}_{\rm cc} = \frac{1}{N}\sum_{k \in (1,N)} 
    \big\| \hat{C}^{out^{\mathbb{S}_k}}_k (p) - \hat{C}^{out^{\mathbb{S}_k}}_k (q)\big\|^{\frac{1}{\mathbb{S}_k}} 
    - 0.1 \cdot \delta_k,
\end{equation}
where $(p,q) \in \{(R,G),(G,B),(R,B)\}$, and
\begin{equation}
   \delta_k = \mathrm{mean}\left(1 - \frac{\min(\hat{C}^{out}_k)}{\mathrm{mean}(\hat{C}^{out}_k)}\right).
\end{equation}
For each view $k$, we learn an adaptive Minkowski parameter $\mathbb{S}_k$ to enable view-specific color correction. The parameter $\mathbb{S}_k$ is predicted using a view-adaptive parameter generator (Fig.~\ref{fig:detail_block}(c)), which shares the architecture of the view-adaptive curve generator. Additionally, a color saturation regularization term $\delta_k$ is introduced to prevent over-desaturation during optimization. Fig.~\ref{fig:color_loss} illustrates the effectiveness of $\mathcal{L}_{\rm cc}$ through visual comparisons and color distribution analysis.

\subsection{Curve Losses}

Beyond image-level objectives, we introduce curve-level constraints on $\mathbb{L}$ to regulate both its values and structural properties. These constraints stabilize pseudo-enhanced image generation and improve overall training robustness.

To avoid excessive deviation of the learned curve, we impose shape regularization using two canonical tone priors: a power curve ${\mathbb{L}}_{po}$ and an S-curve ${\mathbb{L}}_{s}$. These priors guide the learned curve toward plausible tonal transformations while reducing overfitting:
\begin{equation}
\begin{aligned}
 \mathbb{L}_{po}: y &= (x + \epsilon)^{\mathbb G_k}, \quad \epsilon = 10^{-4}, \qquad 0 \leq x \leq 1, \\
 \mathbb{L}_{s}: y &= 
 \begin{cases}
 \mathbb A_k - \mathbb A_k \left( 1 - \frac{x}{\mathbb A_k} \right)^{\mathbb B_k}, & 0 \leq x \leq \mathbb A_k,\\
 \mathbb A_k + (1 - \mathbb A_k)\left( \frac{x - \mathbb A_k}{1 - \mathbb A_k} \right)^{\mathbb B_k}, & \mathbb A_k < x \leq 1,
 \end{cases}
\end{aligned}
\end{equation}
where $\{\mathbb G_k, \mathbb A_k, \mathbb B_k\}$ are view-adaptive learnable parameters predicted by the view-adaptive parameter generator (Fig.~\ref{fig:detail_block}(c)).

To further constrain curve values, we use the cumulative distribution function (CDF) $\mathbb{L}_{\rm cdf}$ derived from histogram equalization of the input image $C^{in}(x)$ as an initialization target. The curve loss is defined as:
\begin{equation}
    \mathcal{L}_{\rm curve} = \omega \|\mathbb{L} - \mathbb{L}_{\rm cdf}\|^2 + 0.5 \|\mathbb{L} - (\mathbb{L}_{po} \cdot \mathbb{L}_{s})\|^2,
\end{equation}
where $\omega$ is set to $1.0$ during the first 3,000 iterations and reduced to $0.1$ thereafter to gradually relax the initialization prior.

Additionally, we impose a total variation (TV) regularization term to encourage smoothness:
\begin{equation}
    \mathcal{L}_{\rm tv} = \frac{1}{255} \sum_{i=0}^{254} |\mathbb{L}(i+1) - \mathbb{L}(i)|^2.
\end{equation}

\subsection{Overall Objective}
The overall training objective of {Luminance-GS++} is:
\begin{equation}
    \mathcal{L}_{\rm total} = \mathcal{L}_{\rm reg} + \mathcal{L}_{\rm spa} + \mathcal{L}_{\rm tv} + 10\,\mathcal{L}_{\rm curve} + \eta\,\mathcal{L}_{\rm cc},
\end{equation}
where $\eta$ is a weighting factor set to $0.1$ when color degradation is present and $0.005$ otherwise.

\begin{table*}[t]
\caption{Experimental results (PSNR $\uparrow$, SSIM $\uparrow$, LPIPS $\downarrow$) on the low-light subset of LOM dataset~\cite{cui_aleth_nerf}. We compare with various enhancement methods~\cite{Zero-DCE,cvpr22_sci,ICCV2023_NeRCo,LLVE_2021_CVPR,SGZ_wacv2022} and NeRF- and 3DGS-based methods~\cite{zou2024enhancing_aaai,cui_aleth_nerf,ye2024gaussian_in_dark}. \colorbox{red!20}{Red} denotes the best result, while \colorbox{yellow!20}{yellow} denotes the second-best.}
\label{tab:LOM_low_full}
\Large
\centering
\renewcommand\arraystretch{1.2}
\begin{adjustbox}{max width = 1.00\linewidth}

\begin{tabular}{ccccccc}
\toprule
\toprule
\multicolumn{1}{c|}{\multirow{2}{*}{Method}} & \multicolumn{1}{c|}{ {\textit{buu}}}                 & \multicolumn{1}{c|}{ {\textit{chair}}}               & \multicolumn{1}{c|}{ {\textit{sofa}}}                 & \multicolumn{1}{c|}{ {\textit{bike}}}                & \multicolumn{1}{c|}{ {\textit{shrub}}}               &  {\textit{mean}}                 \\ \cline{2-7} 
\multicolumn{1}{c|}{}                        & \multicolumn{1}{c|}{PSNR/ SSIM/ LPIPS}   & \multicolumn{1}{c|}{PSNR/ SSIM/ LPIPS}   & \multicolumn{1}{c|}{PSNR/ SSIM/ LPIPS}    & \multicolumn{1}{c|}{PSNR/ SSIM/ LPIPS}   & \multicolumn{1}{c|}{PSNR/ SSIM/ LPIPS}   & PSNR/ SSIM/ LPIPS    \\ \hline
\multicolumn{1}{c|}{3DGS~\cite{3dgs}}   & \multicolumn{1}{c|}{7.53/ 0.299/ 0.442}  & \multicolumn{1}{c|}{6.06/ 0.151/ 0.742}  & \multicolumn{1}{c|}{6.31/ 0.216/ 0.723}  & \multicolumn{1}{c|}{6.37/ 0.077/ 0.781} & \multicolumn{1}{c|}{8.15/ 0.044/ 0.620} &  6.88/ 0.157/ 0.662  \\ \hline
\multicolumn{7}{c}{Image Enhancement Methods + 3DGS}               \\ \hline
\multicolumn{1}{c|}{3DGS + Z-DCE~\cite{Zero-DCE}}         & \multicolumn{1}{c|}{18.02/ 0.834/ 0.303} & \multicolumn{1}{c|}{12.55/ 0.725/ 0.478}  & \multicolumn{1}{c|}{14.66/ 0.822/ 0.460}  & \multicolumn{1}{c|}{10.26/ 0.509/ 0.491}  & \multicolumn{1}{c|}{12.93/ 0.468/ 0.309}  & 13.64/ 0.672/ 0.408 \\
\multicolumn{1}{c|}{Z-DCE~\cite{Zero-DCE} + 3DGS}         & \multicolumn{1}{c|}{17.83/ 0.874/ 0.350} & \multicolumn{1}{c|}{12.47/ 0.762/ 0.399}   & \multicolumn{1}{c|}{13.86/ 0.841/ 0.308} & \multicolumn{1}{c|}{10.37/ 0.544/ 0.441} & \multicolumn{1}{c|}{12.74/ 0.487/ \colorbox{yellow!20}{0.248}} &   13.45/ 0.702/ 0.349 \\
\multicolumn{1}{c|}{3DGS + SCI~\cite{cvpr22_sci}}     & \multicolumn{1}{c|}{13.80/ 0.845/ 0.339}                    & \multicolumn{1}{c|}{19.70/ 0.812/ 0.455}  & \multicolumn{1}{c|}{19.63/ 0.851/ 0.455}    & \multicolumn{1}{c|}{12.86/ 0.621/0.463}   & \multicolumn{1}{c|}{16.14/ 0.600/ 0.442}   & \multicolumn{1}{c}{15.22/ 0.748/ 0.430} \\
\multicolumn{1}{c|}{SCI~\cite{cvpr22_sci} + 3DGS}              & \multicolumn{1}{c|}{7.68/ 0.690/ 0.523}  & \multicolumn{1}{c|}{11.69/ 0.794/ 0.419}  & \multicolumn{1}{c|}{10.02/ 0.770/ 0.365}  & \multicolumn{1}{c|}{13.55/ 0.667/ 0.390} & \multicolumn{1}{c|}{15.72/ 0.538/ 0.339}  & \multicolumn{1}{c}{11.73/ 0.692/ 0.407} \\
\multicolumn{1}{c|}{3DGS + NeRCo~\cite{ICCV2023_NeRCo}}            & \multicolumn{1}{c|}{16.64/ 0.765/ 0.401}      & \multicolumn{1}{c|}{19.24/ 0.759/ 0.466}   & \multicolumn{1}{c|}{16.77/ 0.834/ 0.399}                     & \multicolumn{1}{c|}{16.33/ 0.700/ 0.427}                    & \multicolumn{1}{c|}{17.07/ 0.503/ 0.411}                    &       17.21/ 0.712/ 0.421  \\
\multicolumn{1}{c|}{NeRCo~\cite{ICCV2023_NeRCo} + 3DGS}            & \multicolumn{1}{c|}{16.69/ 0.802/ 0.330}     & \multicolumn{1}{c|}{19.11/ 0.773/ 0.376}  & \multicolumn{1}{c|}{18.04/ 0.868/ 0.381}   & \multicolumn{1}{c|}{16.16/ 0.703/ 0.397}  & \multicolumn{1}{c|}{\colorbox{yellow!20}{17.97}/ 0.502/ 0.399}                    &  17.59/ 0.727/ 0.345  \\ \hline
\multicolumn{7}{c}{Video Enhancement Methods + 3DGS}                                                               \\ \hline
\multicolumn{1}{c|}{LLVE~\cite{LLVE_2021_CVPR} + 3DGS}             & \multicolumn{1}{c|}{19.67/ 0.868/ 0.253} & \multicolumn{1}{c|}{15.29/ 0.805/ 0.424} & \multicolumn{1}{c|}{17.18/ 0.858/ 0.379} & \multicolumn{1}{c|}{14.01/ 0.677/ 0.452} & \multicolumn{1}{c|}{15.98/ 0.430/ 0.488} &   16.43/ 0.728/ 0.399 \\
\multicolumn{1}{c|}{SGZ~\cite{SGZ_wacv2022} + 3DGS}    & \multicolumn{1}{c|}{19.21/ 0.832/ 0.270}    & \multicolumn{1}{c|}{12.30/ 0.755/ 0.377}                & \multicolumn{1}{c|}{14.54/ 0.815/ 0.329}  & \multicolumn{1}{c|}{10.61/ 0.563/ \colorbox{red!20}{0.375}}     & \multicolumn{1}{c|}{14.04/ 0.565/ 0.416}   &   14.14/ 0.706/ 0.353  \\ \hline
\multicolumn{7}{c}{NeRF- and 3DGS-based Enhancement Methods}                                                                              \\ \hline
\multicolumn{1}{c|}{AME-NeRF~\cite{zou2024enhancing_aaai}}               & \multicolumn{1}{c|}{19.89/ 0.854/ 0.312} & \multicolumn{1}{c|}{17.05/ 0.751/ 0.381} & \multicolumn{1}{c|}{17.93/ 0.847/ 0.378}  & \multicolumn{1}{c|}{18.14/ 0.732/ 0.437} & \multicolumn{1}{c|}{15.23/ 0.462/ 0.518} & 17.65/ 0.729/ 0.405  \\
\multicolumn{1}{c|}{Aleth-NeRF~\cite{cui_aleth_nerf}}       & \multicolumn{1}{c|}{\colorbox{yellow!20}{20.22}/ 0.859/ 0.315} & \multicolumn{1}{c|}{\colorbox{yellow!20}{20.93}/ 0.818/ 0.468} & \multicolumn{1}{c|}{19.52/ 0.857/ 0.354}   & \multicolumn{1}{c|}{\colorbox{red!20}{20.46}/ 0.727/ 0.499} & \multicolumn{1}{c|}{\colorbox{red!20}{18.24}/ 0.511/ 0.448} & \colorbox{yellow!20}{19.87}/ 0.754/ 0.417  
\\
\multicolumn{1}{c|}{Gaussian-DK~\cite{ye2024gaussian_in_dark}}       & \multicolumn{1}{c|}{15.95/ 0.753/ 0.247} & \multicolumn{1}{c|}{14.86/ 0.693/ \colorbox{yellow!20}{0.318}} & \multicolumn{1}{c|}{14.39/ 0.782/ 0.350}   & \multicolumn{1}{c|}{17.83/ 0.687/ 0.394} & \multicolumn{1}{c|}{12.47/ 0.138/ 0.565} & 15.10/ 0.610/ 0.375
\\
\hline
\multicolumn{7}{c}{Our Proposed Method}  
\\ \hline

\multicolumn{1}{c|}{ {Luminance-GS~\cite{cui_luminance_gs}}}            & \multicolumn{1}{c|}{18.09/ \colorbox{yellow!20}{0.877}/ \colorbox{yellow!20}{0.193}}                    & \multicolumn{1}{c|}{19.82/ \colorbox{yellow!20}{0.835}/ 0.367}                    & \multicolumn{1}{c|}{\colorbox{yellow!20}{20.12}/ \colorbox{yellow!20}{0.871}/ \colorbox{yellow!20}{0.259}} & \multicolumn{1}{c|}{18.27/ \colorbox{yellow!20}{0.749}/ 0.411}   & \multicolumn{1}{c|}{15.40/ \colorbox{red!20}{0.666}/ \colorbox{red!20}{0.241}} &   18.34/ \colorbox{yellow!20}{0.799}/ \colorbox{yellow!20}{0.294}

\\
\multicolumn{1}{c|}{ {Luminance-GS++}}        & \multicolumn{1}{c|}{\colorbox{red!20}{24.75}/ \colorbox{red!20}{0.924}/ \colorbox{red!20}{0.164}} & \multicolumn{1}{c|}{\colorbox{red!20}{21.95}/ \colorbox{red!20}{0.860}/ \colorbox{red!20}{0.309}} & \multicolumn{1}{c|}{\colorbox{red!20}{23.01}/ \colorbox{red!20}{0.910}/ \colorbox{red!20}{0.242}}  & \multicolumn{1}{c|}{\colorbox{yellow!20}{19.54}/ \colorbox{red!20}{0.764}/ \colorbox{yellow!20}{0.385}} & \multicolumn{1}{c|}{14.45/ \colorbox{yellow!20}{0.656}/ 0.262} & \multicolumn{1}{c}{\colorbox{red!20}{20.74}/ \colorbox{red!20}{0.823}/ \colorbox{red!20}{0.272}}\\
\bottomrule
\bottomrule
\end{tabular}
\end{adjustbox}
\vspace{-4mm}
\end{table*}

\vspace{-1mm}
\section{Experiments}
\label{sec6:exp}
This section presents the experiments validating the effectiveness of {Luminance-GS++}. We first introduce the datasets and the baseline settings, followed by the experimental results under lightness-degradation, color-degradation, and mixed-degradation scenarios. 
Finally, we provide comprehensive ablation studies to examine the contribution of each proposed component.

\subsection{Experimental Settings}

\subsubsection{Datasets}\hfill

\textit{{Lightness Degradation}}: 
To evaluate novel view synthesis (NVS) under lightness degradation, we follow the experimental protocol of our conference version~\cite{cui_luminance_gs} and consider three scenarios: low-light, overexposure, and varying exposure.

For low-light and overexposure settings, we use the LOM benchmark~\cite{cui_aleth_nerf}, which contains five scenes (“{\textit{buu}},” “{\textit{chair}},” “{\textit{sofa}},” “{\textit{bike}},” and “{\textit{shrub}}”) captured using a DJI Osmo Action~3 camera. Each scene includes 25–65 multi-view images with three illumination levels (low-light, normal-light, and overexposure), obtained by varying exposure time and ISO settings. Following~\cite{cui_aleth_nerf}, low-light and overexposed views are used for training, while normal-light images serve as evaluation targets.

\begin{figure*}[tp]
    \centering
    \includegraphics[width=0.92\linewidth]{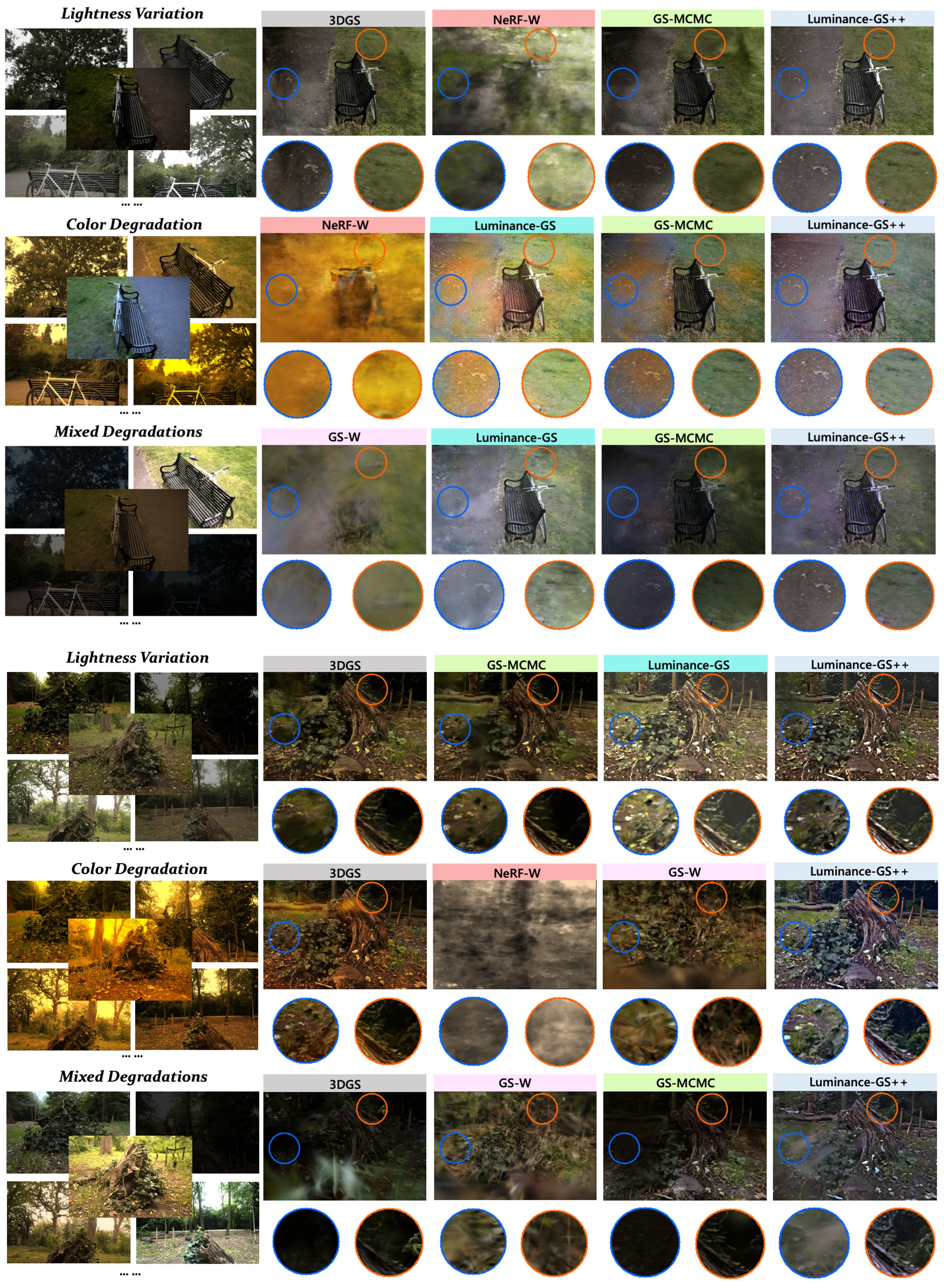}
    \caption{Qualitative comparisons of NVS results under lightness, color, and mixed degradation. The last column presents our Luminance-GS++, while the remaining columns correspond to competing methods~\cite{3dgs,GS-W_ECCV2024,martinbrualla2020nerfw,cui_luminance_gs,MCMC}. Zoomed-in regions  are provided to highlight details.}
    \label{fig:Mix_exp}
\end{figure*}

For lightness variation experiments, we adopt the unbounded MipNeRF360 dataset~\cite{barron2022mipnerf360}, consistent with Luminance-GS~\cite{cui_luminance_gs}. We use seven scenes (“{\textit{bicycle}},” “{\textit{bonsai}},” “{\textit{counter}},” “{\textit{garden}},” “{\textit{kitchen}},” “{\textit{room}},” and “{\textit{stump}}”). Following~\cite{cui_luminance_gs}, varying-exposure inputs are synthesized by applying exposure scaling~\cite{Afifi_2021_CVPR} and gamma perturbations to increase task difficulty. The synthesis process is defined as:
\begin{equation}
\begin{aligned}
    C^{in}(x) = (C^{ori}(x) \cdot K)^{\gamma}, \\
    K \in (0.05, 0.8) \cup (1.25, 3.0), \quad \gamma \in (0.8, 2.5),
\end{aligned}
\end{equation}
where $C^{ori}(x)$ denotes the original MipNeRF360 images and $C^{in}(x)$ denotes the synthesized varying-exposure inputs. Example results are shown in Fig.~\ref{fig:Mix_exp}.

\textit{{Color Degradation}}:
We evaluate robustness to color degradation on the Mip-NeRF360 dataset~\cite{barron2022mipnerf360} by synthesizing realistic illuminant variations. Following the Planckian Jitter framework~\cite{iclr_color_jitter}, we model the illuminant color $\boldsymbol{\rho}_T \in \mathbb{R}^3$ by sampling a black-body radiator temperature $T$ according to Planck’s Law~\cite{planck_law} and mapping the resulting spectrum to sRGB space. Color degradation is applied using a Von Kries transform combined with brightness and contrast perturbations. The degraded image $C^{in}(x)$ is defined as:
\begin{equation}
\begin{aligned}
C^{in}(x) = & \; c_B \cdot c_C \cdot (C^{ori}(x) \odot \boldsymbol{\rho}_T) \\
& + (1 - c_C)\cdot \mu\big(c_B \cdot (C^{ori}(x) \odot \boldsymbol{\rho}_T)\big),
\end{aligned}
\end{equation}
where $\odot$ denotes element-wise multiplication, $c_B$ and $c_C$ respectively represent brightness and contrast factors~\cite{iclr_color_jitter}, and $\mu(\cdot)$ denotes the mean operator.

Following~\cite{iclr_color_jitter}, we evaluate three illumination conditions: (1) Cool lighting ($8000\text{K} \le T \le 9500\text{K}$) for “\textit{bonsai},” “\textit{counter},” and “\textit{garden}”; (2) Warm lighting ($1800\text{K} \le T \le 2500\text{K}$) for “\textit{kitchen},” “\textit{room},” and “\textit{stump}”; and (3) Mixed lighting for “\textit{bicycle},” where views are sampled across a wide temperature range ($1800\text{K}$–$9500\text{K}$).

\begin{figure}[tp]
    \centering
    \includegraphics[width=0.99\linewidth]{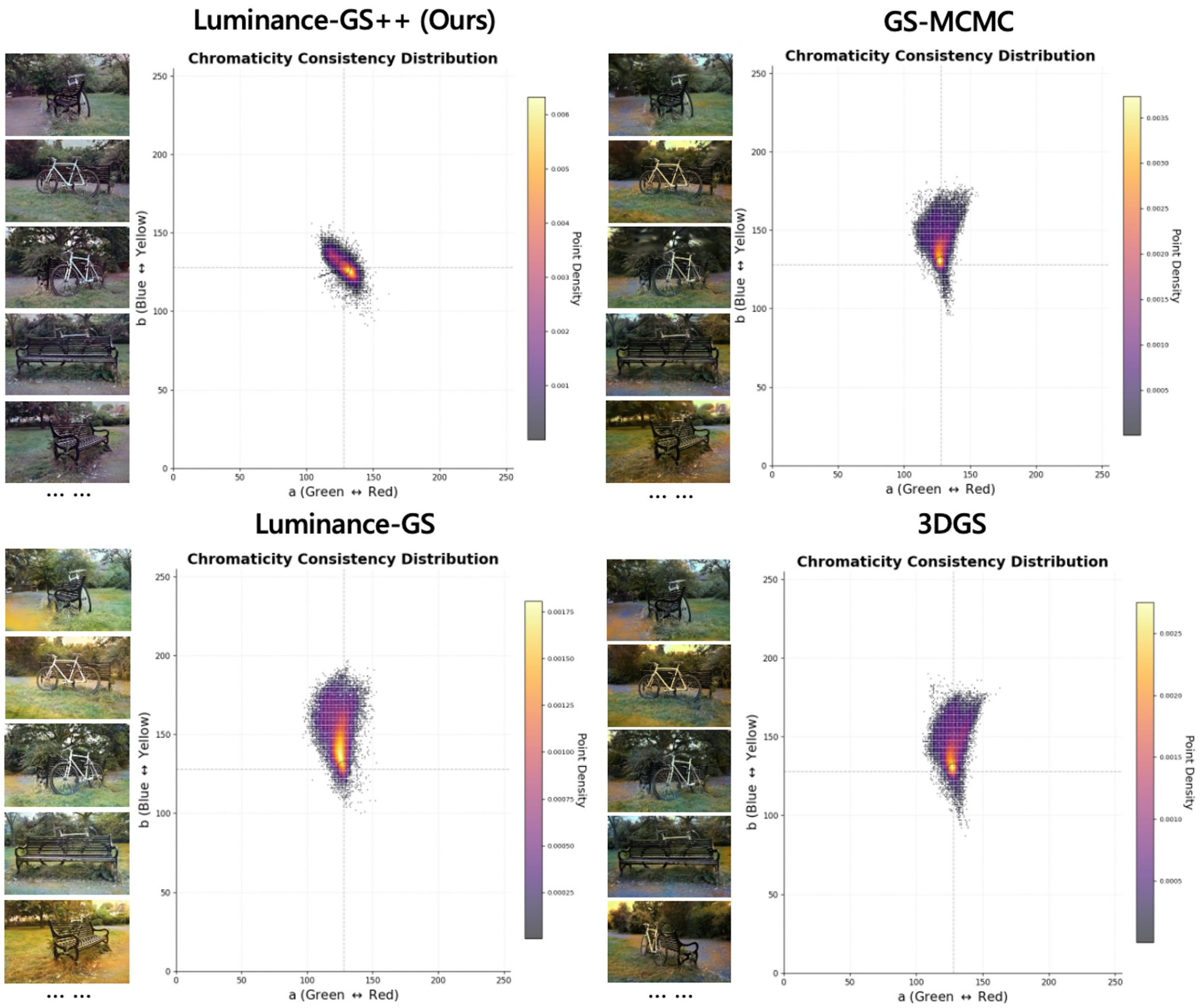}
    \vspace{-3.5mm}
    \caption{Chromaticity consistency visualization in CIELAB color space. 
Pixel chromaticity coordinates ($a^*, b^*$) from rendered views are plotted for a static region. Luminance-GS++ produces a compact and concentrated distribution, indicating improved cross-view color consistency and stable photometric appearance.}
    \vspace{-3.5mm}
    \label{fig:Lab_Visual}
\end{figure}

\textit{{Mixed Degradation}}: 
For this setting, we use the MipNeRF-360 dataset~\cite{barron2022mipnerf360}, where lightness and color degradations are jointly applied to simulate challenging real-world conditions. Such combined degradations across multiple views significantly increase the difficulty of consistent 3D reconstruction. The degraded input is defined as:
\begin{equation}
\begin{aligned}
     C^{in}(x) = & (colordeg(C^{ori}(x)) \cdot K)^{\gamma}, \\
     K  \in & (0.05, 0.8) \cup (1.25, 3.0), \\ 
    T  \in & (2500\text{K}, 8000\text{K}),  
    \gamma \in (0.8, 2.5),
\end{aligned}
\end{equation}
where $colordeg(\cdot)$ denotes the color degradation process described previously. Synthesized examples are shown in Fig.~\ref{fig:Mix_exp}.

\subsubsection{Comparison Methods}\hfill

We compare against two categories of baselines: 
(1) state-of-the-art 2D image/video restoration methods~\cite{Zero-DCE,cvpr22_sci,ICCV2023_NeRCo,LLVE_2021_CVPR,SGZ_wacv2022} applied as preprocessing for 3DGS, and 
(2) recent NeRF- and 3DGS-based methods designed for robust 3D reconstruction under challenging conditions~\cite{zou2024enhancing_aaai,cui_aleth_nerf,ye2024gaussian_in_dark}. 

\vspace{1mm}
\noindent \textbf{Image Restoration Methods} 
target single-view enhancement and therefore do not explicitly enforce coherence or consistency in multi-view 3D settings. We include three state-of-the-art image enhancement methods (Z-DCE~\cite{Zero-DCE}, SCI~\cite{cvpr22_sci}, and NerCo~\cite{ICCV2023_NeRCo}) and three exposure correction methods (MSEC~\cite{Afifi_2021_CVPR}, IAT~\cite{Cui_2022_BMVC}, and MSLT~\cite{zhou2024mslt}). 
Since single-image approaches cannot directly produce multi-view-consistent outputs, we integrate them with 3DGS~\cite{3dgs} using two evaluation pipelines: 
\ding{172} applying restoration as a preprocessing prior to 3DGS reconstruction (denoted as `` ''+3DGS in Tab.~\ref{tab:LOM_low_full}); and 
\ding{173} performing restoration as a post-processing after novel view generation (denoted as 3DGS+`` '' in Tab.~\ref{tab:LOM_low_full}).

\begin{table*}[tp]
\caption{Experimental results (PSNR $\uparrow$, SSIM $\uparrow$, LPIPS $\downarrow$) on the overexposure scenes of LOM~\cite{cui_aleth_nerf} . We compare with exposure correction methods~\cite{Afifi_2021_CVPR,Cui_2022_BMVC,zhou2024mslt} and NeRF-based methods~\cite{cui_aleth_nerf}. \colorbox{red!20}{Red} denotes the best result while \colorbox{yellow!20}{yellow} denotes the second-best.}
\label{tab:LOM_oe_full}
\Large
\centering
\renewcommand\arraystretch{1.2}
\begin{adjustbox}{max width = 1\linewidth}
\begin{tabular}{ccccccc}
\toprule
\toprule
\multicolumn{1}{c|}{\multirow{2}{*}{Method}} & \multicolumn{1}{c|}{{\textit{``buu"}}}                 & \multicolumn{1}{c|}{{\textit{``chair"}}}               & \multicolumn{1}{c|}{{\textit{``sofa"}}}                 & \multicolumn{1}{c|}{{\textit{``bike"}}}                & \multicolumn{1}{c|}{{\textit{``shrub"}}}               & {\textit{mean}}                 \\ \cline{2-7} 
\multicolumn{1}{c|}{}                        & \multicolumn{1}{c|}{PSNR/ SSIM/ LPIPS}   & \multicolumn{1}{c|}{PSNR/ SSIM/ LPIPS}   & \multicolumn{1}{c|}{PSNR/ SSIM/ LPIPS}    & \multicolumn{1}{c|}{PSNR/ SSIM/ LPIPS}   & \multicolumn{1}{c|}{PSNR/ SSIM/ LPIPS}   & PSNR/ SSIM/ LPIPS    \\ \hline
\multicolumn{1}{c|}{3DGS~\cite{3dgs}}   & \multicolumn{1}{c|}{6.96/ 0.674/ 0.609}  & \multicolumn{1}{c|}{11.14/ 0.790/ 0.362}  & \multicolumn{1}{c|}{10.17/ 0.790/ 0.369}  & \multicolumn{1}{c|}{9.58/ 0.730/ 0.323} & \multicolumn{1}{c|}{10.34/ 0.646/ 0.299} &   9.64/ 0.726/ 0.392  \\ \hline
\multicolumn{7}{c}{Exposure Correction Methods + 3DGS}               \\ \hline
\multicolumn{1}{c|}{3DGS + MSEC~\cite{Afifi_2021_CVPR}}         & \multicolumn{1}{c|}{16.03/ 0.806/ 0.517} & \multicolumn{1}{c|}{20.81/ 0.851/ 0.408}  & \multicolumn{1}{c|}{20.65/ 0.862/ 0.397}  & \multicolumn{1}{c|}{22.10/ 0.826/ 0.305}  & \multicolumn{1}{c|}{18.21/ 0.678/ 0.289}  &  19.56/ 0.805/ 0.382  \\
\multicolumn{1}{c|}{MSEC~\cite{Afifi_2021_CVPR} + 3DGS}         & \multicolumn{1}{c|}{15.08/ 0.804/ 0.440} & \multicolumn{1}{c|}{16.63/ 0.797/ 0.416}   & \multicolumn{1}{c|}{20.09/ 0.828/ 0.335} & \multicolumn{1}{c|}{17.57/ 0.739/ 0.368} & \multicolumn{1}{c|}{16.61/ 0.666/ 0.255} &  17.20/ 0.767/ 0.363  \\
\multicolumn{1}{c|}{3DGS + IAT~\cite{Cui_2022_BMVC}}    & \multicolumn{1}{c|}{15.34/ 0.804/ 0.522}    & \multicolumn{1}{c|}{21.96/ 0.833/ 0.292}                    & \multicolumn{1}{c|}{20.23/ 0.872/ 0.402}                     & \multicolumn{1}{c|}{22.36/ 0.832/ 0.291}                    & \multicolumn{1}{c|}{\colorbox{red!20}{21.24}/ 0.765/ 0.226}                    & \multicolumn{1}{c}{20.23/ 0.821/ 0.347} \\
\multicolumn{1}{c|}{IAT~\cite{Cui_2022_BMVC} + 3DGS}  & \multicolumn{1}{c|}{15.86/ 0.803/ 0.387}  & \multicolumn{1}{c|}{18.61/ 0.830/ 0.367}  & \multicolumn{1}{c|}{17.42/ 0.833/ 0.348} & \multicolumn{1}{c|}{19.17/ 0.801/ 0.235}  & \multicolumn{1}{c|}{16.74/ 0.731/ 0.219} & 17.56/ 0.800/ 0.311 \\
\multicolumn{1}{c|}{3DGS + MSLT~\cite{zhou2024mslt}}            & \multicolumn{1}{c|}{15.34/ 0.798/ 0.473}    & \multicolumn{1}{c|}{21.69/ 0.823/ 0.304}                    & \multicolumn{1}{c|}{\colorbox{red!20}{23.05}/ 0.830/ 0.317}                     & \multicolumn{1}{c|}{23.37/ 0.830/ 0.317}                    & \multicolumn{1}{c|}{\colorbox{yellow!20}{18.89}/ 0.779/ 0.214}                    &   \multicolumn{1}{c}{20.39/ 0.815/ 0.345}  \\
\multicolumn{1}{c|}{MSLT~\cite{zhou2024mslt} + 3DGS}            & \multicolumn{1}{c|}{16.35/ 0.805/ 0.333}     & \multicolumn{1}{c|}{20.93/ 0.828/ 0.275}  & \multicolumn{1}{c|}{\colorbox{yellow!20}{21.65}/ 0.847/ 0.259}   & \multicolumn{1}{c|}{24.03/ 0.841/ 0.244}  & \multicolumn{1}{c|}{18.29/ \colorbox{red!20}{0.797}/ 0.199}   &  \multicolumn{1}{c}{20.25/ 0.824/ 0.262} \\ \hline

\multicolumn{7}{c}{NeRF-based Exposure Correction Method}                                             \\ \hline

\multicolumn{1}{c|}{Aleth-NeRF~\cite{cui_aleth_nerf}}       & \multicolumn{1}{c|}{16.78/ 0.805/ 0.611} & \multicolumn{1}{c|}{20.08/ 0.820/ 0.499} & \multicolumn{1}{c|}{17.85/ 0.852/ 0.458}   & \multicolumn{1}{c|}{19.85/ 0.773/ 0.392} & \multicolumn{1}{c|}{15.91/ 0.477/ 0.483} & 18.09/ 0.745/ 0.488  \\
 \hline
\multicolumn{7}{c}{Our Proposed Method}                                                                      \\ \hline

\multicolumn{1}{c|}{{Luminance-GS~\cite{cui_luminance_gs}}}            & \multicolumn{1}{c|}{\colorbox{yellow!20}{19.67}/ \colorbox{yellow!20}{0.811}/ \colorbox{yellow!20}{0.311}}                    & \multicolumn{1}{c|}{\colorbox{yellow!20}{22.63}/ \colorbox{yellow!20}{0.856}/ \colorbox{red!20}{0.207}}                    & \multicolumn{1}{c|}{21.16/ \colorbox{yellow!20}{0.878}/ \colorbox{red!20}{0.204}}                     & \multicolumn{1}{c|}{\colorbox{yellow!20}{24.05}/ \colorbox{yellow!20}{0.851}/ \colorbox{yellow!20}{0.216}}                    & \multicolumn{1}{c|}{16.04/ \colorbox{yellow!20}{0.780}/ \colorbox{yellow!20}{0.173}}   & \colorbox{yellow!20}{20.71}/ \colorbox{yellow!20}{0.835}/ \colorbox{red!20}{0.222}    \\ 

\multicolumn{1}{c|}{{Luminance-GS++}}            & \multicolumn{1}{c|}{\colorbox{red!20}{19.86}/ \colorbox{red!20}{0.878}/ \colorbox{red!20}{0.304}}                    & \multicolumn{1}{c|}{\colorbox{red!20}{24.61}/ \colorbox{red!20}{0.888}/ \colorbox{yellow!20}{0.237}}                    & \multicolumn{1}{c|}{21.54/ \colorbox{red!20}{0.889}/ \colorbox{yellow!20}{0.216}}                     & \multicolumn{1}{c|}{\colorbox{red!20}{25.13}/ \colorbox{red!20}{0.854}/ \colorbox{red!20}{0.192}}                    & \multicolumn{1}{c|}{15.92/ 0.762/ \colorbox{red!20}{0.166}}   & \colorbox{red!20}{21.41}/ \colorbox{red!20}{0.854}/ \colorbox{yellow!20}{0.223} 
\\
\bottomrule
\bottomrule
\end{tabular}
\end{adjustbox}
\vspace{-3mm}
\end{table*}
\begin{table}
\caption{Training time comparisons on the LOM~\cite{cui_aleth_nerf} and Mip-NeRF360~\cite{barron2022mipnerf360} datasets.}
\vspace{-2mm}
\label{tab:Train_Time}
\Huge
\centering
\renewcommand\arraystretch{1.2}
\begin{adjustbox}{max width = 1\linewidth}
\begin{tabular}{c|cccccc}
\toprule
\toprule
Methods                                                                   & Aleth-NeRF & AME-NeRF & 3DGS & Gaussian-DK & Luminance-GS & Luminance-GS++ \\ \hline
\begin{tabular}[c]{@{}c@{}}LOM\\ (GPU hours)\end{tabular}         & 14.08      & 11.28    & 0.08 & 0.58        & 0.13         & 0.23           \\ \hline
Methods                                                                   & NeRF-W     & 3DGS     & GS-W & GS-MCMC     & Luminance-GS & Luminance-GS++ \\ \hline
\begin{tabular}[c]{@{}c@{}}Mip-NeRF360\\ (GPU hours)\end{tabular} & 32.67      & 0.10     & 0.58 & 0.14        & 0.20         & 0.28           \\ 
\bottomrule
\bottomrule
\end{tabular}
\end{adjustbox}
\vspace{-3mm}
\end{table}

\vspace{1mm}
\noindent \textbf{Video Restoration Methods} 
emphasize temporal consistency across frames, leading to improved spatial and temporal coherence compared to single-image approaches. We include two state-of-the-art video enhancement methods, LLVE~\cite{LLVE_2021_CVPR} and SGZ~\cite{SGZ_wacv2022}, for comparison. Since these methods are designed to enhance existing views rather than synthesize novel ones, we integrate them with 3DGS~\cite{3dgs} to ensure a fair comparison within the NVS pipeline.

\begin{figure}[t]
    \centering
    \includegraphics[width=0.92\linewidth]{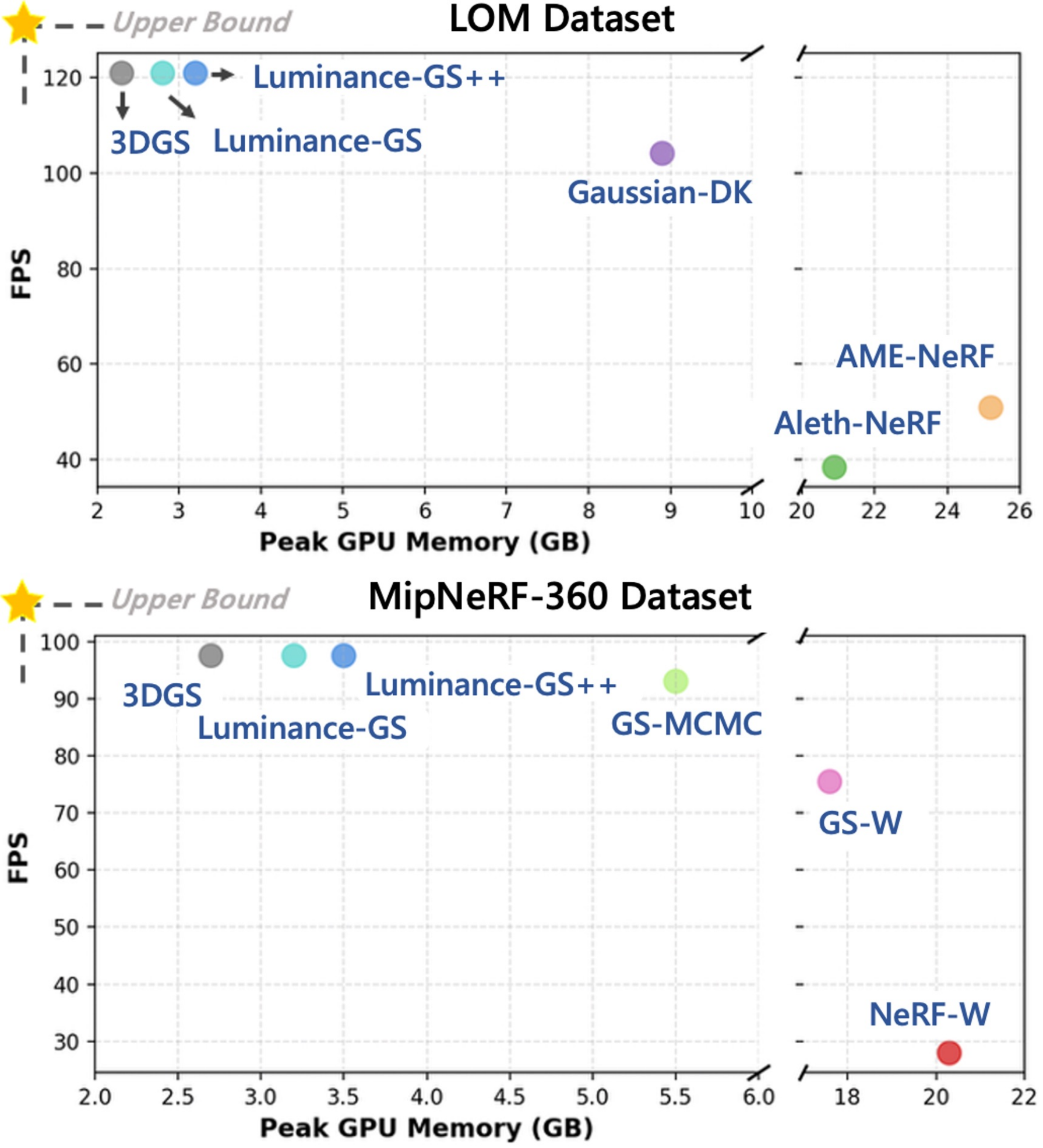}
    \vspace{-1.5mm}
    \caption{Efficiency comparison: FPS and peak GPU memory (GB) on the LOM~\cite{cui_aleth_nerf} and Mip-NeRF 360~\cite{barron2022mipnerf360} datasets. Methods closer to the upper-left corner indicate better efficiency.}
    \vspace{-3.5mm}
    \label{fig:Efficiency}
\end{figure}

\vspace{1mm}
\noindent \textbf{Novel View Synthesis (NVS) Methods}. 
We compare with a range of state-of-the-art NVS approaches: 
{\ding{172} Aleth-NeRF}~\cite{cui_aleth_nerf}, a NeRF-based method for low-light and overexposed scenes based on the ``Concealing Fields'' hypothesis; 
{\ding{173} AME-NeRF}~\cite{zou2024enhancing_aaai}, which addresses low-light reconstruction using a dual-level optimization framework; 
{\ding{174} Gaussian-DK}~\cite{ye2024gaussian_in_dark}, which integrates 3DGS with learnable tone mapping for low-light view synthesis; 
{\ding{175} NeRF-W}~\cite{martinbrualla2020nerfw}, which introduces learnable appearance embeddings and transient modeling for view synthesis; 
{\ding{176} GS-W}~\cite{GS-W_ECCV2024}, a 3DGS-based extension of NeRF-W with dynamic appearance modeling; 
{\ding{177} GS-MCMC}~\cite{MCMC}, which reformulates 3DGS densification and pruning as MCMC-based state transitions; 
and {\ding{178} Luminance-GS}~\cite{cui_luminance_gs}, the conference version of our method.

\begin{figure*}[tp]
    \centering
    \includegraphics[width=0.92\linewidth]{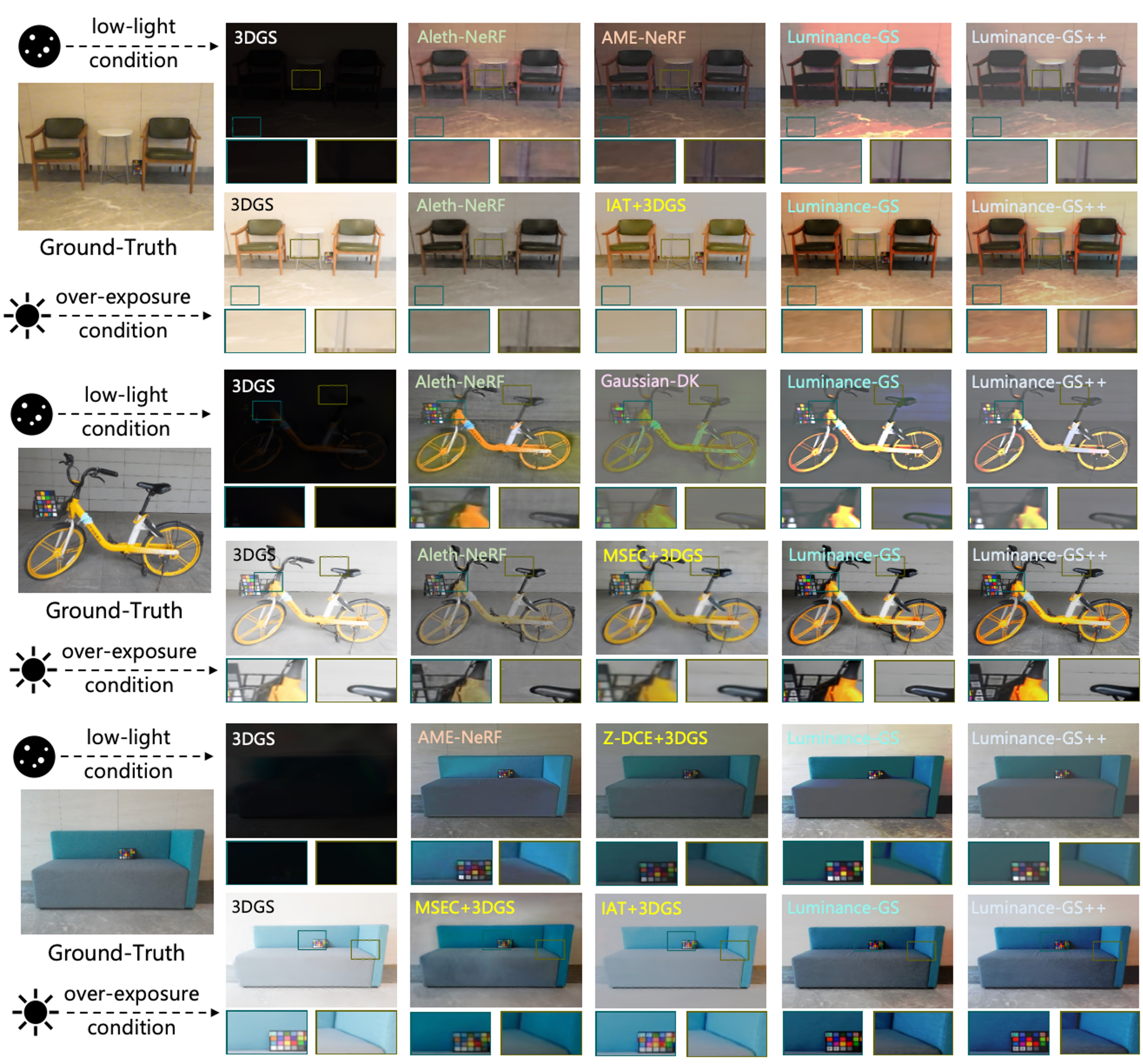}
    \vspace{-1mm} 
    \caption{Experimental results on the low-light and overexposure subsets of LOM dataset~\cite{cui_aleth_nerf}. The first column shows the results of the basic 3DGS~\cite{3dgs}, while the second and third columns present the results of various comparison methods~\cite{cui_aleth_nerf,zou2024enhancing_aaai,ye2024gaussian_in_dark,Cui_2022_BMVC,Afifi_2021_CVPR,Zero-DCE}. The last two columns show the results of our {Luminance-GS}~\cite{cui_luminance_gs} and {Luminance-GS++}. Some regions are zoomed in to better observe the details.}
    \label{fig:LOM_exp}
    \vspace{-4mm} 
\end{figure*}

\begin{table*}[tp]
\caption{Experimental results (PSNR ↑, SSIM ↑, LPIPS ↓) on the MipNeRF-360~\cite{barron2022mipnerf360} dataset under three degradation settings: lightness variation, color degradation, and mixed degradation. In the color degradation setting, \includegraphics[height=1.2em]{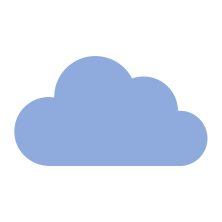} denotes cool color temperature, while \includegraphics[height=1.2em]{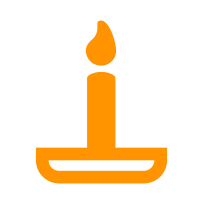} denotes warm color temperature.}
\vspace{-2mm}
\label{tab:Exp_all}
\large
\centering
\renewcommand\arraystretch{1.2}
\begin{adjustbox}{max width = 1\linewidth}
\begin{tabular}{c|llllll}
\toprule \toprule
\multirow{2}{*}{{\textit{Lightness Variation}}} & \multicolumn{1}{c}{3DGS~\cite{3dgs}} & \multicolumn{1}{c}{NeRF-W~\cite{martinbrualla2020nerfw}} & \multicolumn{1}{c}{GS-W~\cite{GS-W_ECCV2024}} & \multicolumn{1}{c}{GS-MCMC~\cite{MCMC}} & \multicolumn{1}{c}{{Luminance-GS}~\cite{cui_luminance_gs}} & \multicolumn{1}{c}{{Luminance-GS++}} \\ \cline{2-7}   & PSNR/ SSIM/ LPIPS        &  PSNR/ SSIM/ LPIPS  & PSNR/ SSIM/ LPIPS   & PSNR/ SSIM/ LPIPS   &  PSNR/ SSIM/ LPIPS    &           PSNR/ SSIM/ LPIPS     \\ \hline
{\textit{bicycle}}              & \colorbox{yellow!20}{18.63}/0.539/0.409        & 13.82/0.233/0.631          & 18.33/0.618/0.377        & 18.47/0.591/0.343           & 18.52/\colorbox{yellow!20}{0.647}/\colorbox{yellow!20}{0.334}                & \colorbox{red!20}{20.93}/\colorbox{red!20}{0.678}/\colorbox{red!20}{0.302}                  \\
{\textit{bonsai}}               & 12.50/0.299/0.560        & 9.64/0.232/0.725           & 15.46/0.541/\colorbox{yellow!20}{0.421}        & 13.28/0.371/0.470           & \colorbox{yellow!20}{15.61}/\colorbox{yellow!20}{0.560}/0.438                & \colorbox{red!20}{16.13}/\colorbox{red!20}{0.612}/\colorbox{red!20}{0.340}                  \\
{\textit{counter}}              & 14.01/0.491/0.376        & 10.81/0.411/0.758          & 15.79/0.618/0.344        & 15.64/0.584/\colorbox{red!20}{0.249}           & 16.29/\colorbox{yellow!20}{0.631}/\colorbox{yellow!20}{0.313}                & \colorbox{red!20}{18.01}/\colorbox{red!20}{0.677}/\colorbox{yellow!20}{0.289}                  \\
{\textit{garden}}               & 20.39/0.755/0.205        & 10.71/0.222/0.678          & \colorbox{yellow!20}{20.49}/0.765/0.222        & 19.24/0.768/\colorbox{yellow!20}{0.186}           & \colorbox{red!20}{20.93}/\colorbox{yellow!20}{0.786}/0.193                & 19.76/\colorbox{red!20}{0.790}/\colorbox{red!20}{0.179}                  \\
{\textit{kitchen}}              & \colorbox{yellow!20}{22.26}/0.832/0.177        & 13.79/0.502/0.601          & 20.77/0.802/0.167        & 20.87/\colorbox{yellow!20}{0.840}/\colorbox{yellow!20}{0.109}           & \colorbox{red!20}{23.09}/0.834/0.144                & 21.81/\colorbox{red!20}{0.845}/\colorbox{red!20}{0.099}                  \\
{\textit{room}}                 & 14.22/0.549/\colorbox{yellow!20}{0.299}        & 12.89/0.432/0.534          & \colorbox{red!20}{17.19}/0.558/0.378        & 15.12/0.605/0.326           & 16.77/\colorbox{yellow!20}{0.656}/0.320                & \colorbox{yellow!20}{17.02}/\colorbox{red!20}{0.675}/\colorbox{red!20}{0.298}                  \\
{\textit{stump}}                & 16.61/\colorbox{yellow!20}{0.532}/0.388        & 13.98/0.403/0.712          & 14.65/0.503/0.429        & \colorbox{yellow!20}{18.86}/0.499/\colorbox{yellow!20}{0.379}           & 15.63/\colorbox{yellow!20}{0.532}/0.388                & \colorbox{red!20}{18.88}/\colorbox{red!20}{0.591}/\colorbox{red!20}{0.368}                  \\
{\textit{mean}}                 & 16.95/0.571/0.345        & 12.23/0.348/0.663          & 17.52/0.629/0.334        & 17.35/0.608/\colorbox{yellow!20}{0.295}           & \colorbox{yellow!20}{18.12}/\colorbox{yellow!20}{0.664}/0.304                & \colorbox{red!20}{18.93}/\colorbox{red!20}{0.696}/\colorbox{red!20}{0.268}                  \\ \midrule \midrule
\multirow{2}{*}{{\textit{Color Degradation}}} & \multicolumn{1}{c}{3DGS~\cite{3dgs}} & \multicolumn{1}{c}{NeRF-W~\cite{martinbrualla2020nerfw}} & \multicolumn{1}{c}{GS-W~\cite{GS-W_ECCV2024}} & \multicolumn{1}{c}{GS-MCMC~\cite{MCMC}} & \multicolumn{1}{c}{{Luminance-GS}~\cite{cui_luminance_gs}} & \multicolumn{1}{c}{{Luminance-GS++}} \\ \cline{2-7}   & PSNR/ SSIM/ LPIPS        &  PSNR/ SSIM/ LPIPS  & PSNR/ SSIM/ LPIPS   & PSNR/ SSIM/ LPIPS   &  PSNR/ SSIM/ LPIPS    &           PSNR/ SSIM/ LPIPS     \\ \hline
{\textit{bicycle}} \includegraphics[height=1.2em]{pics/icon_cloud.png} \includegraphics[height=1.2em]{pics/icon_candle.png}             & \colorbox{yellow!20}{19.57}/\colorbox{yellow!20}{0.700}/0.357        &  11.21/0.273/0.812   &  16.39/0.356/0.837   & 18.26/0.629/0.463           &   14.45/0.665/\colorbox{yellow!20}{0.328}   & \colorbox{red!20}{19.67}/\colorbox{red!20}{0.730}/\colorbox{red!20}{0.284}                  \\
{\textit{bonsai}}  \includegraphics[height=1.2em]{pics/icon_cloud.png}               & \colorbox{red!20}{18.43}/\colorbox{yellow!20}{0.808}/\colorbox{yellow!20}{0.169}        &  13.98/0.398/0.771   &  15.33/0.495/0.670  & 17.12/0.785/0.224           &  13.15/0.666/0.212 & \colorbox{yellow!20}{17.45}/\colorbox{red!20}{0.821}/\colorbox{red!20}{0.157}                  \\
{\textit{counter}} \includegraphics[height=1.2em]{pics/icon_cloud.png}             & 17.21/0.704/0.239        & 14.51/0.309/0.728  &   13.95/0.428/0.637    & \colorbox{red!20}{18.00}/\colorbox{yellow!20}{0.722}/0.331           &   13.62/0.673/\colorbox{red!20}{0.225}   & \colorbox{yellow!20}{17.22}/\colorbox{red!20}{0.756}/\colorbox{yellow!20}{0.234}                  \\
{\textit{garden}}  \includegraphics[height=1.2em]{pics/icon_cloud.png}             & 18.74/\colorbox{red!20}{0.828}/\colorbox{red!20}{0.221}        &   17.52/0.299/0.835   &  17.20/0.365/0.832  & \colorbox{yellow!20}{18.78}/\colorbox{yellow!20}{0.823}/0.261           &  16.67/0.810/\colorbox{yellow!20}{0.226}   & \colorbox{red!20}{20.24}/\colorbox{red!20}{0.828}/0.234                  \\
{\textit{kitchen}} \includegraphics[height=1.2em]{pics/icon_candle.png}              & 13.41/\colorbox{yellow!20}{0.739}/\colorbox{red!20}{0.305}        &  15.68/0.417/0.706  & \colorbox{yellow!20}{16.37}/0.422/0.595   & 13.68/0.721/0.382           &   12.00/0.638/0.406   & \colorbox{red!20}{17.92}/\colorbox{red!20}{0.749}/\colorbox{yellow!20}{0.350}                  \\
{\textit{room}}  \includegraphics[height=1.2em]{pics/icon_candle.png}                & 16.88/\colorbox{yellow!20}{0.745}/0.335        &   14.18/0.392/0.788    &  13.68/0.548/0.680   & \colorbox{yellow!20}{16.98}/0.733/0.310           &   13.11/0.668/\colorbox{red!20}{0.252}    & \colorbox{red!20}{17.19}/\colorbox{red!20}{0.758}/\colorbox{yellow!20}{0.288}                  \\
{\textit{stump}} \includegraphics[height=1.2em]{pics/icon_candle.png}               & \colorbox{yellow!20}{18.56}/0.606/0.422        &   15.23/0.322/0.801  &  17.66/0.338/0.728  & 18.53/\colorbox{yellow!20}{0.611}/\colorbox{yellow!20}{0.383}           &   10.64/0.538/0.448   & \colorbox{red!20}{19.19}/\colorbox{red!20}{0.617}/\colorbox{red!20}{0.374}                  \\
{\textit{mean}}                 & \colorbox{yellow!20}{17.97}/\colorbox{yellow!20}{0.747}/\colorbox{yellow!20}{0.295}        &  14.62/0.344/0.777  &   15.80/0.422/0.711   & 17.33/0.718/0.336           &  13.38/0.665/0.300   & \colorbox{red!20}{18.41}/\colorbox{red!20}{0.751}/\colorbox{red!20}{0.274}                  \\ \midrule \midrule
\multirow{2}{*}{{\textit{Mixed Degradation}}} & \multicolumn{1}{c}{3DGS~\cite{3dgs}} & \multicolumn{1}{c}{NeRF-W~\cite{martinbrualla2020nerfw}} & \multicolumn{1}{c}{GS-W~\cite{GS-W_ECCV2024}} & \multicolumn{1}{c}{GS-MCMC~\cite{MCMC}} & \multicolumn{1}{c}{{Luminance-GS}~\cite{cui_luminance_gs}} & \multicolumn{1}{c}{{Luminance-GS++}} \\ \cline{2-7}   & PSNR/ SSIM/ LPIPS        &  PSNR/ SSIM/ LPIPS  & PSNR/ SSIM/ LPIPS   & PSNR/ SSIM/ LPIPS   &  PSNR/ SSIM/ LPIPS    &           PSNR/ SSIM/ LPIPS     \\ \hline

{\textit{bicycle}}              & 14.12/0.350/0.631        & 9.65/0.176/0.654           & \colorbox{yellow!20}{16.13}/0.355/0.896        & 13.60/0.401/0.522           & 16.03/\colorbox{yellow!20}{0.542}/\colorbox{yellow!20}{0.469}                & \colorbox{red!20}{19.56}/\colorbox{red!20}{0.577}/\colorbox{red!20}{0.466}                  \\
{\textit{bonsai}}               & \colorbox{yellow!20}{16.32}/0.632/0.280        & 10.19/0.198/0.732          & 14.79/0.476/0.415        & 14.92/0.545/0.334           & 15.45/\colorbox{yellow!20}{0.665}/\colorbox{yellow!20}{0.233}                & \colorbox{red!20}{18.16}/\colorbox{red!20}{0.728}/\colorbox{red!20}{0.228}                  \\
{\textit{counter}}              & \colorbox{yellow!20}{16.13}/0.617/0.355        & 9.88/0.375/0.722           & 14.27/0.438/0.557        & 14.58/0.527/0.320           & 16.01/\colorbox{yellow!20}{0.649}/\colorbox{yellow!20}{0.315}                & \colorbox{red!20}{16.94}/\colorbox{red!20}{0.660}/\colorbox{red!20}{0.304}                  \\
{\textit{garden}}               & 15.55/\colorbox{yellow!20}{0.606}/\colorbox{red!20}{0.374}        & 13.98/0.216/0.694          & 15.87/0.364/0.561        & 14.12/0.561/0.399           & \colorbox{yellow!20}{16.14}/0.585/0.402                & \colorbox{red!20}{16.46}/\colorbox{red!20}{0.647}/\colorbox{yellow!20}{0.389}                  \\
{\textit{kitchen}}              & 17.87/\colorbox{yellow!20}{0.731}/0.280        & 7.52/0.508/0.675           & 15.47/0.644/0.393        & 18.07/0.722/\colorbox{yellow!20}{0.261}           & \colorbox{red!20}{18.22}/0.688/0.295                & \colorbox{yellow!20}{18.14}/\colorbox{red!20}{0.738}/\colorbox{red!20}{0.257}                  \\
{\textit{room}}                 & 15.60/\colorbox{yellow!20}{0.665}/0.331        & 11.19/0.398/0.677          & 13.63/0.551/0.497        & \colorbox{yellow!20}{15.67}/0.607/0.335           & \colorbox{yellow!20}{15.67}/0.661/\colorbox{red!20}{0.314}                & \colorbox{red!20}{17.10}/\colorbox{red!20}{0.705}/\colorbox{yellow!20}{0.317}                  \\
{\textit{stump}}                & 15.48/0.328/0.562        & 10.29/0.378/0.722          & \colorbox{yellow!20}{17.40}/0.340/0.709        & 14.51/0.377/0.577           & 15.30/\colorbox{yellow!20}{0.500}/\colorbox{red!20}{0.427}                & \colorbox{red!20}{18.67}/\colorbox{red!20}{0.554}/\colorbox{yellow!20}{0.453}                  \\
{\textit{mean}}                 & 15.87/0.561/0.402        & 10.39/0.321/0.697          & 15.37/0.453/0.575        & 15.07/0.534/0.393           & \colorbox{yellow!20}{16.12}/\colorbox{yellow!20}{0.613}/\colorbox{red!20}{0.350}                & \colorbox{red!20}{17.86}/\colorbox{red!20}{0.658}/\colorbox{yellow!20}{0.352}                  \\ \bottomrule \bottomrule
\end{tabular}
\end{adjustbox}
\vspace{-3mm}
\end{table*}

\subsection{Lightness Degradation Experiments}

Tables~\ref{tab:LOM_low_full} and~\ref{tab:LOM_oe_full} present quantitative comparisons on the LOM dataset~\cite{cui_aleth_nerf} under low-light and overexposure conditions. In low-light scenarios (Tab.~\ref{tab:LOM_low_full}), image enhancement methods generally achieve limited performance, primarily due to the lack of multi-view consistency and weak zero-shot generalization. Among them, NeRCo~\cite{ICCV2023_NeRCo} outperforms Z-DCE~\cite{Zero-DCE} and SCI~\cite{cvpr22_sci}. Video enhancement methods~\cite{LLVE_2021_CVPR,SGZ_wacv2022} achieve slightly improved LPIPS scores compared to image enhancement and NeRF-based approaches, suggesting that temporal modeling provides better cross-view coherence.

For NVS-specific methods, prior works such as AME-NeRF~\cite{zou2024enhancing_aaai}, Aleth-NeRF~\cite{cui_aleth_nerf}, and Gaussian-DK~\cite{ye2024gaussian_in_dark} often exhibit color shifts or geometric artifacts (see Fig.~\ref{fig:LOM_exp}). In contrast, Luminance-GS++ achieves the best overall performance (highlighted in red in Table~\ref{tab:LOM_low_full}) across nearly all scene-specific metrics and the overall mean. Compared with Luminance-GS~\cite{cui_luminance_gs}, our method improves mean PSNR by 13.1\% (18.34\,dB $\rightarrow$ 20.74\,dB), demonstrating the effectiveness of the proposed local refinement branch. Qualitative results (Fig.~\ref{fig:LOM_exp}) further confirm improved restoration of fine-grained details.

A similar trend is observed under overexposure (Tab.~\ref{tab:LOM_oe_full}). Aleth-NeRF~\cite{cui_aleth_nerf} and exposure correction methods~\cite{zhou2024mslt,Afifi_2021_CVPR,Cui_2022_BMVC} exhibit inferior LPIPS performance, although MSLT~\cite{zhou2024mslt} achieves relatively stronger PSNR and SSIM among exposure-based methods. Luminance-GS++ consistently achieves the best overall results and surpasses Luminance-GS across all evaluation metrics.

For lightness variation experiments (Tab.~\ref{tab:Exp_all}), we compare against 3DGS~\cite{3dgs}, in-the-wild methods (NeRF-W~\cite{martinbrualla2020nerfw}, GS-W~\cite{GS-W_ECCV2024}), and the robust NVS method GS-MCMC~\cite{MCMC}. NeRF-W and GS-W generally produce suboptimal results; for example, NeRF-W introduces blurry artifacts in 360-degree scenes (Fig.~\ref{fig:Mix_exp}). Although GS-MCMC achieves competitive LPIPS scores, it suffers from abrupt lighting transitions and inconsistencies. In contrast, Luminance-GS++ consistently outperforms all baselines across nearly all scenes, demonstrating strong robustness to lightness variations. As illustrated in the ``bicycle'' scene (Fig.~\ref{fig:Mix_exp}), our method maintains spatially uniform illumination, particularly in low-texture regions such as the ground.

\subsection{Color Degradation Experiments}

For this setting, we conduct comprehensive validations across cool lighting, warm lighting, and mixed lighting conditions (Tab.~\ref{tab:Exp_all}). Overall, Luminance-GS++ consistently outperforms the baseline methods, achieving either the best or second-best performance across most evaluation metrics.

Beyond standard quantitative metrics, we further analyze chromaticity consistency across rendered views. As shown in Fig.~\ref{fig:Lab_Visual}, rendered images are projected into the CIELab color space, where discrepancies in pixel-wise chromaticity coordinates ($a^*, b^*$) are measured across viewpoints. Existing baselines exhibit dispersed chromaticity distributions, indicating inconsistent color alignment. In contrast, Luminance-GS++ produces compact and concentrated chromaticity clusters, demonstrating improved cross-view color constancy. Additional qualitative comparisons in Fig.~\ref{fig:Lab_Visual} and Fig.~\ref{fig:Mix_exp} further highlight the effectiveness of our method in preserving consistent color tones under challenging illumination variations.

\subsection{Mixed Degradation Experiments}

Quantitative comparisons under mixed degradation are reported in Tab.~\ref{tab:Exp_all}. Luminance-GS++ consistently achieves strong performance, obtaining the best results across most evaluation metrics and scenarios. Qualitative comparisons are shown in Fig.~\ref{fig:Mix_exp}. 

Under challenging mixed-degradation conditions, the in-the-wild method GS-W~\cite{GS-W_ECCV2024} struggles to identify a stable reference appearance, leading to inconsistent and visually disordered renderings. GS-MCMC~\cite{MCMC} produces noticeable non-uniform illumination artifacts, indicating limited robustness to jointly varying lightness and color degradations. In contrast, Luminance-GS++ generates visually coherent results with spatially consistent brightness and well-balanced color reproduction across views.

\begin{table*}[tp]
\caption{Ablation analysis of each module, including the global curve ${\mathbb L}^g$, curve bias ${\mathbb L}^b_k$, per-view matrix $\mathcal{M}_k$, and the residual branch ${\mathbb R}$.}
\label{tab:Exp_ablation}
\large
\vspace{-3mm}
\centering
\renewcommand\arraystretch{1.1}
\begin{adjustbox}{max width = 1\linewidth}
\begin{tabular}{cccc|ccccc}
\toprule \toprule
${\mathbb L}^g$ & ${\mathbb L}^b_k$ & $\mathcal{M}_k$ & ${\mathbb R}$ & \begin{tabular}[c]{@{}c@{}}LOM~\cite{cui_aleth_nerf} \\ (low-light) {\textit{buu}}\end{tabular} & \begin{tabular}[c]{@{}c@{}}LOM~\cite{cui_aleth_nerf} \\ (overexposure) {\textit{buu}}\end{tabular} & \begin{tabular}[c]{@{}c@{}}MipNeRF360~\cite{barron2022mipnerf360} \\ (lightness variation) {\textit{bicycle}}\end{tabular} & \begin{tabular}[c]{@{}c@{}}MipNeRF360~\cite{barron2022mipnerf360} \\ (color degradation) {\textit{``bicycle''}}\end{tabular} & \begin{tabular}[c]{@{}c@{}}MipNeRF360~\cite{barron2022mipnerf360} \\ (mixed degradation) {\textit{``bicycle''}}\end{tabular} \\ \midrule \midrule
$\checkmark$  &    &    &   &  17.96/ 0.797/ 0.210  & 19.03/ 0.788/ 0.345 & 14.08/ 0.493/ 0.432  &  13.76/ 0.601/ 0.412  &  12.87/ 0.471/ 0.579 \\ \hline
   & $\checkmark$  &   &  & 17.79/ 0.654/ 0.375 & 16.45/ 0.713/ 0.462 & 17.03/ 0.584/ 0.397 & 13.65/ 0.599/ 0.427  & 15.42/ 0.488/ 0.556 \\ \hline
$\checkmark$  & 
$\checkmark$  &    &    & 18.03/ 0.858/ 0.199 &  19.35/ 0.803/ 0.346 & 17.01/ 0.592/ 0.388 & 14.05/ 0.629/ 0.355   &  15.78/ 0.507/ 0.548  \\ \hline 
$\checkmark$  & 
$\checkmark$  & 
$\checkmark$ &   & 18.11/ 0.879/ 0.192  & 19.69/ 0.812/ 0.313 &  16.94/ 0.606/ 0.384  & 14.01/ 0.650/ 0.309  &  15.96/ 0.502/ 0.530  \\ \hline
   &  &  & $\checkmark$  & 10.86/ 0.691/ 0.409  & 17.86/ 0.752/ 0.413  &  19.51/ 0.631/ 0.369  &  {21.07}/ 0.719/ 0.289 &  18.27/ 0.531/ 0.508 \\ \hline
$\checkmark$  &    & $\checkmark$  & $\checkmark$   & 24.53/ 0.911/ 0.165  &  19.77/ {0.878}/ 0.312 & 19.96/ 0.648/ 0.366 & 19.82/ 0.724/ 0.286  & 18.74/ 0.548/ 0.494 \\ \hline
   & $\checkmark$  & $\checkmark$  & $\checkmark$   & 19.21/ 0.901/ 0.179  & 17.95/ 0.842/ 0.354  & 20.41/ 0.663/ 0.323 &  19.54/ 0.724/ 0.291 &     19.12/ 0.563/ 0.468 \\ \hline
$\checkmark$  & $\checkmark$  & $\checkmark$  & $\checkmark$  & {24.75}/ {0.924}/ {0.164} & {19.86}/ {0.878}/ {0.304} & {20.93}/ {0.678}/ {0.302} & 19.67/ {0.730}/ {0.284} & {19.56}/ {0.577}/ {0.466} \\ \bottomrule \bottomrule
\end{tabular}
\end{adjustbox}
\vspace{-3mm}
\end{table*}

\subsection{Efficiency Evaluation}

We evaluate computational efficiency by comparing model training time (GPU hours), as summarized in Tab.~\ref{tab:Train_Time}. The results demonstrate that Luminance-GS++ reduces training costs by orders of magnitude compared to NeRF-based methods (e.g., Aleth-NeRF~\cite{cui_aleth_nerf} and NeRF-W~\cite{martinbrualla2020nerfw}). Despite introducing additional components to enhance reconstruction quality, our approach maintains training efficiency comparable to other 3DGS-based variants.

\begin{table}[tp]
\caption{Ablation analysis of different loss functions.}
\label{tab:ablation_loss}
\vspace{-3mm}
\centering
\large
\renewcommand\arraystretch{1.5}
\begin{adjustbox}{max width = 1\linewidth}
\begin{tabular}{c|cccc}
\hline \hline
    & w/o $\mathcal{L}_{\rm spa}$  & w/o $\mathcal{L}_{\rm curve}$ & w/o  $\mathcal{L}_{\rm cc}$ & Full \\ \hline
LOM (low-light) {\textit{buu}} & 13.05/ 0.809/ 0.322  & 14.48/ 0.789/ 0.459  & 24.69/ 0.918/ {0.162} & {24.75}/ {0.924}/ 0.164  \\ \hline
\begin{tabular}[c]{@{}c@{}}MipNeRF360 \\ (color degradation) {\textit{bicycle}}\end{tabular} & {21.11}/ 0.710/ 0.324  & 19.57/ 0.612/ 0.449 & 20.35/ 0.728/ 0.295 & 19.67/ {0.730}/ {0.284}  \\ \hline
\begin{tabular}[c]{@{}c@{}}MipNeRF360 \\ (mixed degradation) {\textit{bicycle}}\end{tabular} & 17.17/ 0.466/ 0.569 & 15.08/ 0.387/ 0.657  & 19.39/ 0.550/ 0.496 & {19.56}/ {0.577}/ {0.466}  \\ \hline \hline
\end{tabular}
\end{adjustbox}
\vspace{-4mm}
\end{table}

\begin{table}[tp]
\caption{Ablation analysis of different residual branch blocks.}
\label{tab:ablation_block}
\vspace{-3mm}
\large
\centering
\renewcommand\arraystretch{1.4}
\begin{adjustbox}{max width = 1\linewidth}
\begin{tabular}{l|ccc}
\hline \hline
               & LOM (low-light) & LOM (overexposure) & MipNeRF-360 (mixed degradation) \\ \hline
ResNet~\cite{resnet} Block   & {20.77}/ 0.812/ 0.279     & 21.33/ 0.851/ 0.226  &  {17.88}/ 0.651/ 0.367 \\ \hline
PVT~\cite{pvt} Block   & 19.98/ 0.811/ 0.301  & 20.82/ 0.836/ 0.233     & 17.12/ 0.631/ 0.389                \\ \hline
ConvNext~\cite{liu2022convnet} Block &  20.74/ {0.823}/ {0.272}  &  {21.41}/ {0.854}/ {0.223}   & 17.86/ {0.658}/ {0.352}  \\ \hline \hline
\end{tabular}
\end{adjustbox}
\vspace{-4mm}
\end{table}

Furthermore, we evaluate runtime efficiency in terms of rendering FPS and GPU memory consumption. As illustrated in Fig.~\ref{fig:Efficiency}, Luminance-GS++ occupies the upper-left region of the performance plot, indicating a favorable trade-off between rendering speed and memory efficiency. Our method achieves real-time rendering performance comparable to the original 3DGS while requiring substantially less GPU memory than NeRF-based approaches, as well as Gaussian-DK~\cite{ye2024gaussian_in_dark} and GS-W~\cite{GS-W_ECCV2024}.

\begin{figure}[tp]
    \centering
    \includegraphics[width=0.98\linewidth]
    {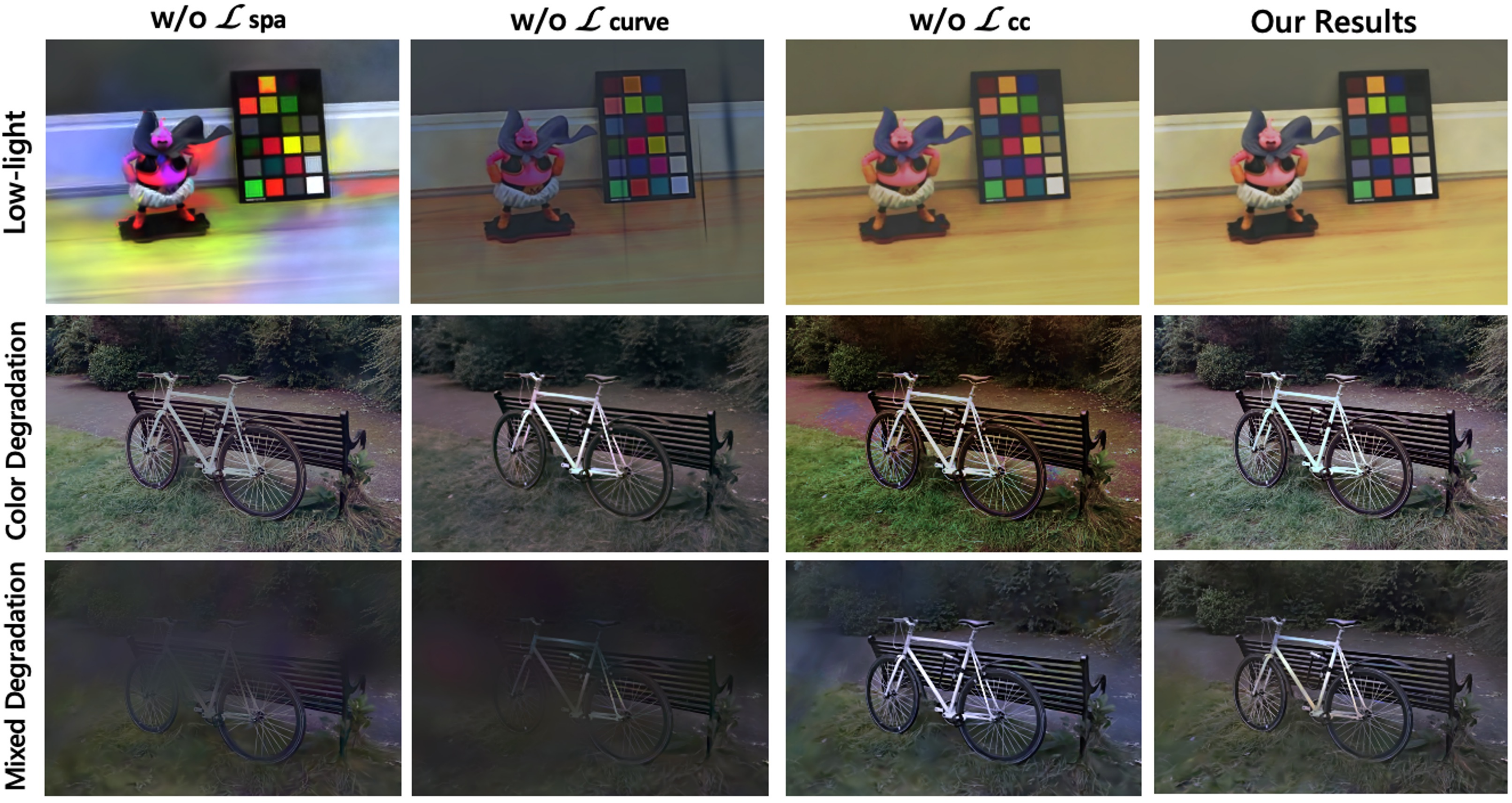}
    \vspace{-2mm}
    \caption{Visualization of the effectiveness of different loss components.}
    \label{fig:Loss_ablation}
    \vspace{-4mm}
\end{figure}

\subsection{Ablation Studies}

We conduct comprehensive ablation studies to systematically evaluate the contribution of each proposed component and loss function across diverse scenarios.

We first analyze the impact of individual modules in Tab.~\ref{tab:ablation_block}. The full model consistently achieves the best performance, confirming the necessity of each component. In particular, the residual branch $\mathbb{R}$ is critical for handling complex degradations; for example, under ``mixed degradation,'' incorporating $\mathbb{R}$ improves PSNR from 15.96\,dB to 19.56\,dB. However, $\mathbb{R}$ relies on structural guidance from the global curve $\mathbb{L}^g$. When applied in isolation on the low-light dataset, performance collapses to 10.86\,dB, whereas the full configuration restores performance to 24.75\,dB. This indicates that $\mathbb{L}^g$ provides a stable luminance baseline, while $\mathbb{R}$ refines high-frequency details and color deviations.

We further evaluate the contribution of each loss component in Tab.~\ref{tab:ablation_loss}. All loss terms contribute positively to performance, with the spatial consistency loss $\mathcal{L}_{\rm spa}$ playing a particularly important role under illumination degradations. Qualitative comparisons in Fig.~\ref{fig:Loss_ablation} further highlight the benefits of individual losses: the curve loss $\mathcal{L}_{\rm curve}$ improves overall visual quality, while the color correction loss $\mathcal{L}_{\rm cc}$ enhances cross-view color consistency. Additional examples are provided in Fig.~\ref{fig:color_loss}.

Finally, we analyze architectural choices within the residual branch $\mathbb{R}$ by replacing the ConvNeXt~\cite{liu2022convnet} block with alternatives including ResNet~\cite{resnet} and PVT~\cite{pvt} (Tab.~\ref{tab:ablation_block}). ConvNeXt yields the best performance, suggesting that strong local modeling capability is essential for effective residual correction.

\section{Conclusion}
We present {Luminance-GS++}, a framework that improves the robustness of 3D Gaussian Splatting (3DGS) under challenging illumination and color variations. Without modifying the explicit representation of 3DGS, our method formulates illumination correction as a pseudo-label optimization problem that integrates global adjustment and local refinement. We also introduce effective unsupervised objectives to jointly constrain pseudo-label generation and 3D Gaussian attribute learning. Compared with existing NeRF- and 3DGS-based approaches, Luminance-GS++ demonstrates stronger illumination generalizability and achieves state-of-the-art performance across both real-world and synthetic benchmarks.

Looking forward, several directions may further improve reconstruction quality and efficiency. First, enhancing the quality of pseudo-labels remains critical, as improved supervision directly benefits rendering performance. Recent advances in generative models, particularly diffusion-based approaches, offer promising opportunities for high-fidelity pseudo-label generation; however, integrating such models without incurring significant computational overhead remains an open challenge. Second, combining our framework with efficient feed-forward NVS methods~\cite{chen2024mvsplat,jiang2025anysplat} could improve scalability by leveraging large-scale pretraining while avoiding costly per-scene optimization.


\bibliographystyle{IEEEtran}
\bibliography{references}

\vfill

\end{document}